%% file: arxiv.tex
\PassOptionsToPackage{dvipsnames}{xcolor}
\documentclass[10pt,twocolumn,letterpaper]{article}

\input{preamble.tex}

\usepackage[pagebackref=true,breaklinks=true,colorlinks,bookmarks=false]{hyperref}

\iccvfinalcopy 

\begin{document}
\setboolean{inSubfiles}{false}
\setboolean{arxiv}{true}

\subfile{sections/main_title}

\subfile{sections/main_sections}

{\small
\bibliographystyle{ieee_fullname}
\bibliography{egbib}
}

\clearpage

\subfile{sections/supp_sections}

\end{document}

%% file: preamble.tex
\usepackage{caption}
\usepackage{iccv}
\usepackage{times}
\usepackage{epsfig}
\usepackage{graphicx}
\usepackage{amsmath}
\usepackage{amssymb}

\usepackage{graphicx}
\usepackage{amsmath}
\usepackage{amssymb}
\usepackage{booktabs}

\usepackage{enumitem}
\usepackage[dvipsnames]{xcolor}
\usepackage{mathtools}
\usepackage[linesnumbered,ruled,vlined]{algorithm2e}
\usepackage{floatrow}
\usepackage{multirow}
\usepackage{makecell}
\usepackage{etoolbox}
\usepackage{stfloats}

\graphicspath{{resources/}}

\SetKw{KwAnd}{and}

\SetCommentSty{mycommfont}
\SetAlgoNlRelativeSize{0}

\let\originalleft\left
\let\originalright\right
\renewcommand{\left}{\mathopen{}\mathclose\bgroup\originalleft}
\renewcommand{\right}{\aftergroup\egroup\originalright}

\let\originalpartial\partial
\renewcommand{\partial}{\mathop{}\!\mathrm{\originalpartial}}
\newcommand{\expect}{\mathop{\mathbb{E}}}

\newcommand{\ndagger}{\makebox[0pt][l]{\textdagger}}
\newcommand{\nsection}{\makebox[0pt][l]{\textsection}}
\newcommand{\nddagger}{\makebox[0pt][l]{\textdaggerdbl}}
\newcommand{\nast}{\makebox[0pt][l]{\textasteriskcentered}}
\newcommand{\pmplaceholder}{\phantom{\scriptsize\textpm0.00}}

\makeatletter
\renewcommand{\paragraph}{
  \@startsection{paragraph}{4}
  {\z@}{1.0ex \@plus 0ex \@minus .4ex}{-1em}
  {\normalfont\normalsize\bfseries}
}
\makeatother

\makeatletter
\patchcmd{\@algocf@start}
  {-1.5em}
  {-0.5em}
  {}{}
\makeatother

\newboolean{inSubfiles}
\setboolean{inSubfiles}{true} 
\def\bib{\ifthenelse{\boolean{inSubfiles}}
        {\small\bibliographystyle{ieee_fullname}\bibliography{egbib}}
        {}
    }

\newboolean{arxiv}
\setboolean{arxiv}{false} 

\def\iccvPaperID{10495} 

\usepackage{subfiles} 

%% file: sections/main_title.tex
\title{Single-Stage Diffusion NeRF: A Unified Approach to \\ 3D Generation and Reconstruction}

\author{Hansheng Chen,\negmedspace\textsuperscript{1,\textasteriskcentered}
Jiatao Gu,\negmedspace\textsuperscript{2}
Anpei Chen,\negmedspace\textsuperscript{3}
Wei Tian,\negmedspace\textsuperscript{1}
Zhuowen Tu,\negmedspace\textsuperscript{4}
Lingjie Liu,\negmedspace\textsuperscript{5}
Hao Su\textsuperscript{4} \vspace{1ex} \\
\textsuperscript{1}Tongji University \qquad \textsuperscript{2}Apple \qquad \textsuperscript{3}ETH Zürich \qquad \\
\textsuperscript{4}University of California, San Diego \qquad \textsuperscript{5}University of Pennsylvania
}

\twocolumn[{
\maketitle
\begin{center}
    \vspace{-0.4cm}
    \includegraphics[width=0.95\linewidth]{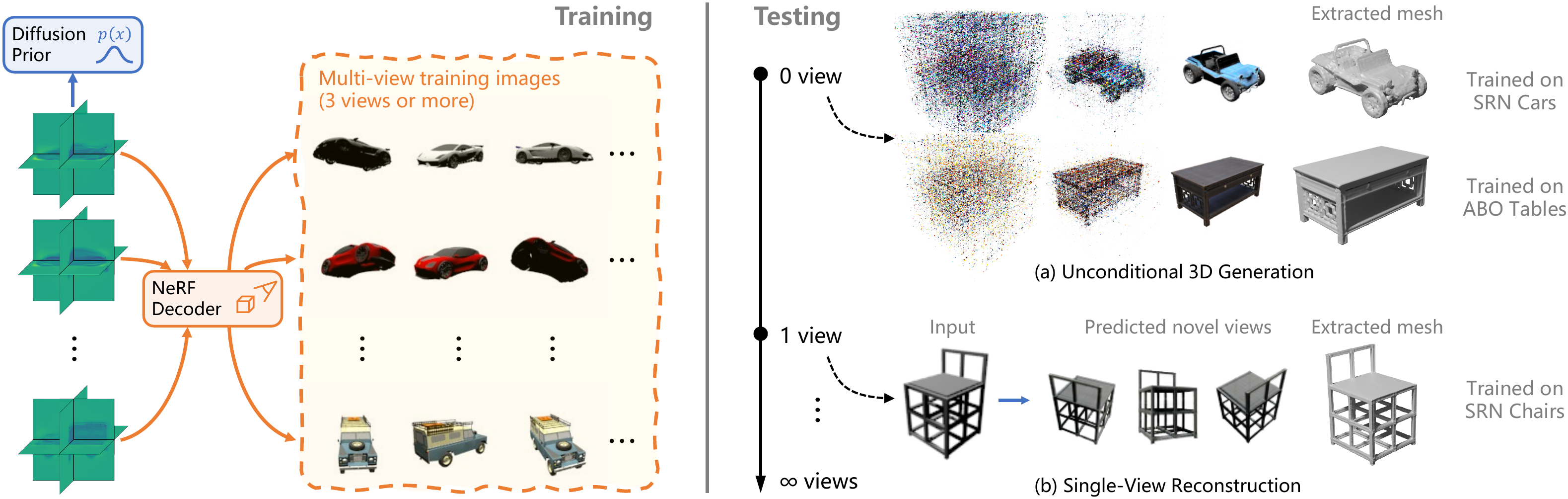}
    \captionof{figure}{During training, SSDNeRF jointly learns triplane features of individual scenes, a shared NeRF decoder, and a triplane diffusion prior. During testing, it can perform (a) unconditional generation, (b) single-view reconstruction, as well as multi-view reconstruction.}
    \label{fig:teaser}
    \vspace{0.3cm}
\end{center}
}]

\ificcvfinal

\begingroup
\renewcommand{\thefootnote}{\fnsymbol{footnote}}
\footnotetext[1]{Work done during a remote internship with UCSD.}
\endgroup

\setcounter{footnote}{5}

\fi

%% file: sections/main_sections.tex
\begin{abstract}

3D-aware image synthesis encompasses a variety of tasks, such as scene generation and novel view synthesis from images. 
Despite numerous task-specific methods, developing a comprehensive model remains challenging.
In this paper, we present SSDNeRF, a unified approach that employs an expressive diffusion model to learn a generalizable prior of neural radiance fields (NeRF) from multi-view images of diverse objects. 
Previous studies have used two-stage approaches that rely on pretrained NeRFs as real data to train diffusion models. 
In contrast, we propose a new single-stage training paradigm with an end-to-end objective that jointly optimizes a NeRF auto-decoder and a latent diffusion model, enabling simultaneous 3D reconstruction and prior learning, even from sparsely available views. 
At test time, we can directly sample the diffusion prior for unconditional generation, or combine it with arbitrary observations of unseen objects for NeRF reconstruction. SSDNeRF demonstrates robust results comparable to or better than leading task-specific methods in unconditional generation and single/sparse-view 3D reconstruction.\ificcvfinal\footnote{Project page: \url{https://lakonik.github.io/ssdnerf}}\fi
\ificcvfinal\vspace{-0.5cm}\fi

\end{abstract}

\section{Introduction}

Synthesizing 3D visual contents has gained significant attention in computer vision and graphics, thanks to advances in neural rendering and generative models. Although numerous methods have emerged to handle individual tasks, such as single-/multi-view 3D reconstruction and 3D content generation, it remains a major challenge to develop a comprehensive framework that bridges the state of the art of multiple tasks. For instance, neural radiance fields (NeRF)~\cite{nerf} have shown impressive results in novel view synthesis by solving the inverse rendering problem via per-scene fitting, which is suitable for dense-view inputs but difficult to generalize to sparse observations. In contrast, many sparse-view 3D reconstruction methods~\cite{pixelnerf, mvsnerf, visionnerf} rely on feed-forward image-to-3D encoders, but they are unable to handle ambiguity in the occluded region and generate crisp images. Regarding unconditional generation, 3D-aware generative adversarial networks (GAN)~\cite{GIRAFFE, eg3d, stylenerf, get3d} are partially limited in their usage of single-image discriminators, which cannot reason cross-view relationships to effectively learn from multi-view data.

In this paper, we propose a unified approach to various 3D tasks (Fig.~\ref{fig:teaser}) by developing a holistic model that learns generalizable 3D priors from multi-view images. Inspired by the success of 2D diffusion models~\cite{ddpm, ddim, sdedit, latentdiffusion, repaint}, we present the \textbf{S}ingle-\textbf{S}tage \textbf{D}iffusion \textbf{NeRF} (SSDNeRF), which models the generative prior of scene latent codes with a 3D latent diffusion model (LDM). 

While similar LDMs have been applied in 2D and 3D generation in previous work~\cite{lsgm,latentdiffusion, functa,gaudi,triplanediff,diffrf}, they typically require two-stage training, where the first stage pretrains the variational auto-encoders (VAE)~\cite{vae} or auto-decoders~\cite{deepsdf} without diffusion models. In the case of diffusion NeRFs, however, we argue that two-stage training induces noisy patterns and artifacts in the latent code due to the uncertain nature of inverse rendering, particularly when training from sparse-view data, which prevents the diffusion model from learning a clean latent manifold effectively.
To address this issue, we introduce a novel single-stage training paradigm that enables end-to-end learning of diffusion and NeRF weights (\textsection~\ref{sstraining}). This approach blends the generative and the rendering biases coherently for improved performance overall and allows for training on sparse-view data. Additionally, we show that the learned 3D priors of unconditional diffusion models can be exploited for flexible test-time scene sampling from arbitrary observations (\textsection~\ref{testmethod}).

We evaluate SSDNeRF on multiple datasets of categorical single-object scenes, demonstrating strong performance overall. Our approach represents a significant step towards a unified framework for various 3D tasks.

To summarize, our main contributions are as follows:
\begin{itemize}[noitemsep,topsep=0.7ex,partopsep=0.7ex]
\item We introduce SSDNeRF, a unified approach to all-round performance in unconditional 3D generation and image-based reconstruction; 
\item We propose a novel single-stage training paradigm that jointly learns NeRF reconstruction and diffusion model from multi-view images of a large number of objects. Notably, this enables training on as sparse as three views per scene, which is previously infeasible;
\item A guidance-finetuning sampling scheme is developed to exploit the learned diffusion priors for 3D reconstruction from arbitrary number of views at test time.
\end{itemize}

\section{Related Work}

\paragraph{3D GANs}
The generative adversarial framework~\cite{gan} has been successfully adapted for 3D generation by integrating projection-based rendering into the generator. A variety of 3D representations have been explored previously, including point clouds, cuboids, spheres~\cite{Liao2020CVPR} and voxels~\cite{HoloGAN} in early works, the more recent radiance fields~\cite{piGAN, graf, devries2021unconstrained, voxgraf, epigraf} and feature fields~\cite{GIRAFFE,stylenerf,eg3d} with volume renderer, and differentiable surface~\cite{get3d} with mesh renderer. 
The above methods are all trained with 2D image discriminators that are unable to reason cross-view relationships, making them heavily dependent on model bias for 3D consistency. As a result, multi-view data cannot be effectively exploited to learn complex and diverse geometries.
3D GANs are mostly applied in unconditional generation. Although 3D completion from images is possible through GAN inversion~\cite{devries2021unconstrained}, faithfulness is not guaranteed due to limited latent expressiveness, as shown in \cite{diffrf, renderdiffusion}.

\paragraph{View-Conditioned Regression and Generation}
Sparse-view 3D reconstruction can be tackled by regressing novel views from input images. Various architectures~\cite{mvsnerf, pixelnerf, visionnerf, nerfusion} have been proposed to encode images into volume features, which can be projected to supervised target views through volume rendering. However, they cannot reason ambiguity and generate diverse and meaningful contents, which often leads to blurry results. 
In contrast, image-conditioned generative models are better at synthesizing distinct contents. 3DiM~\cite{3dim} proposes to generate novel views from a view-conditioned image diffusion model, but the model lacks 3D consistency bias.
\cite{zhou2022sparsefusion, nerdi, nerfdiff} distill priors of image-conditioned 2D diffusion models into NeRFs to enforce 3D constraints. These methods are parallel to our track as they model cross-view relationships in the image space, while our model is inherently 3D.

\paragraph{Auto-Decoders and Diffusion NeRF}
NeRF's per-scene fitting scheme can be generalized to multi-scene fitting by sharing part of the parameters across all scenes, leaving the rest as individual scene codes~\cite{chen2023factor}. Therefore, multi-scene NeRFs can be trained as auto-decoders~\cite{deepsdf}, where the code bank and shared decoder weights are jointly learned. With proper architectures, scene codes can be treated as latents with Gaussian priors, allowing 3D completion and even generation~\cite{codenerf, srn, lolnerf}. However, like 3D GANs, the latents are not expressive enough for faithful reconstruction of detailed objects.
\cite{gaudi, functa, rodin} improve upon vanilla auto-decoders with latent diffusion priors. DiffRF~\cite{diffrf} leverages the diffusion prior to perform 3D completion. These methods train the auto-decoders and diffusion models in two separate stages, which is subject to the limitations in \textsection~\ref{ldmsection}. 

\section{Background}

\subsection{NeRF as an Auto-Decoder}

Given a set of 2D images of a scene and their camera parameters, one can fit a scene model to reconstruct the light field in 3D space, expressed by a plenoptic function $y_\psi(r)$, where $r$ parameterizes the endpoint and direction of a ray in the world space, $\psi$ denotes the scene model parameters, and $y \in \mathbb{R}^3_+$ represents the received light in RGB format.
NeRF~\cite{nerf} represents the light field as integrated radiance along rays through the 3D volume. It models the scene geometry and appearance as functions of the position $p \in \mathbb{R}^3$ and view direction $d \in \mathbb{R}^3$ of a point in the world space, expressed as $\rho_\psi(p)$ and $c_\psi(p, d)$ respectively, where $\rho \in \mathbb{R}_+$ is the density output and $c \in \mathbb{R}^3_+$ is the RGB color output. Differentiable volume rendering is applied to compose the received light $y$ from multiple point samples along a ray $r$.

NeRF can also generalize to multi-scene settings by sharing part of the model parameters across all scenes~\cite{chen2023factor}. Given observations of multiples scenes $\{y_{ij}^\mathrm{gt}, r_{ij}^\mathrm{gt}\}$, where $y_{ij}^\mathrm{gt}, r_{ij}^\mathrm{gt}$ is the $j$-th pair of pixel RGB and ray of the $i$-th scene, one can optimize the per-scene codes $\{x_i\}$ and shared parameters $\psi$ by minimizing the L2 rendering loss:
\begin{equation}
    \mathcal{L}_\mathrm{rend}(\{x_i\},\psi) = \expect_{i}\left[\smash[b]{\sum_{j}}{\frac{1}{2}\left\|y^\mathrm{gt}_{ij} - y_\psi\left(x_i, r^\mathrm{gt}_{ij}\right)\right\|^2}\right].
    \label{rendloss}
\end{equation}
With this objective, the model is trained as an auto-decoder~\cite{deepsdf}, where the scene codes $\{x_i\}$ can be interpreted as the latent codes, and the plenoptic function can be regarded as a decoder in the form of $p_\psi(\{y_j\}|x,\{r_j\}) \coloneqq \prod_j{\mathcal{N}(y_j|y_\psi(x, r_j), I)}$, assuming independent Gaussians.

\paragraph{Challenges in Bridging Generation and Reconstruction}
An auto-decoder with trained weights $\psi$ can perform unconditional generation by decoding latent codes drawn from a Gaussian prior~\cite{lolnerf}. However, to ensure continuity in generation, a low-dimensional latent space and a complex decoder is required, which adds to the difficulty in optimizing the latent code to faithfully reconstruct any given views.

\subsection{Latent Diffusion Models}
\label{ldmsection}

Latent diffusion models (LDM) learn a prior distribution $p_\phi(x)$ in the latent space with parameters $\phi$, which enables the usage of more expressive latent representations, such as 2D grids for images~\cite{lsgm,latentdiffusion}. For neural field generation, previous work~\cite{gaudi, diffrf, functa, triplanediff} adopts a two-stage training scheme, where the auto-decoder is trained first to obtain the per-scene latent $x_i$, which is then treated as real data to train the LDM. The LDM injects Gaussian perturbation $\epsilon \sim \mathcal{N}(0, I)$ into the code $x_i$, yielding a noisy code $x_i^{(t)} \coloneqq \alpha^{(t)} x_i + \sigma^{(t)} \epsilon$ at diffusion time step $t$, under empirical noise schedule functions $\alpha^{(t)}, \sigma^{(t)}$. A denoising network with trainable weights $\phi$ is then tasked with removing the noise from $x_i^{(t)}$ to predict a denoised code $\hat{x}_i$. The network is typically trained with a simplified L2 denoising loss:
\begin{equation}
    \mathcal{L}_\mathrm{diff}(\phi) = \expect_{i,t,\epsilon}{\left[ \frac{1}{2} w^{(t)} \left\| \hat{x}_\phi \left(x^{(t)}_i, t\right) - x_i \right\|^2 \right]},
\label{diffloss}
\end{equation}
where $t\sim\mathcal{U}(0,T)$, $w^{(t)}$ is an empirical time dependent weighting function, and $\hat{x}_\phi(x^{(t)}_i, t)$ formulates the time-conditioned denoising network.

\paragraph{Unconditional/Guided Sampling}

With trained weights $\phi$, one can sample from the diffusion prior $p_\psi(x)$ using a variety of solvers (\eg, DDIM~\cite{ddim}) that recursively denoise $x^{(t)}$, starting from random Gaussian noise $x^{(T)}$, until reaching the denoised state $x^{(0)}$. Moreover, the sampling process can be guided by the gradients of the rendering loss against known observations, allowing 3D reconstruction from images at test time~\cite{diffrf}.  
 
\begin{figure}[t]
\begin{center}
\includegraphics[width=0.84\linewidth]{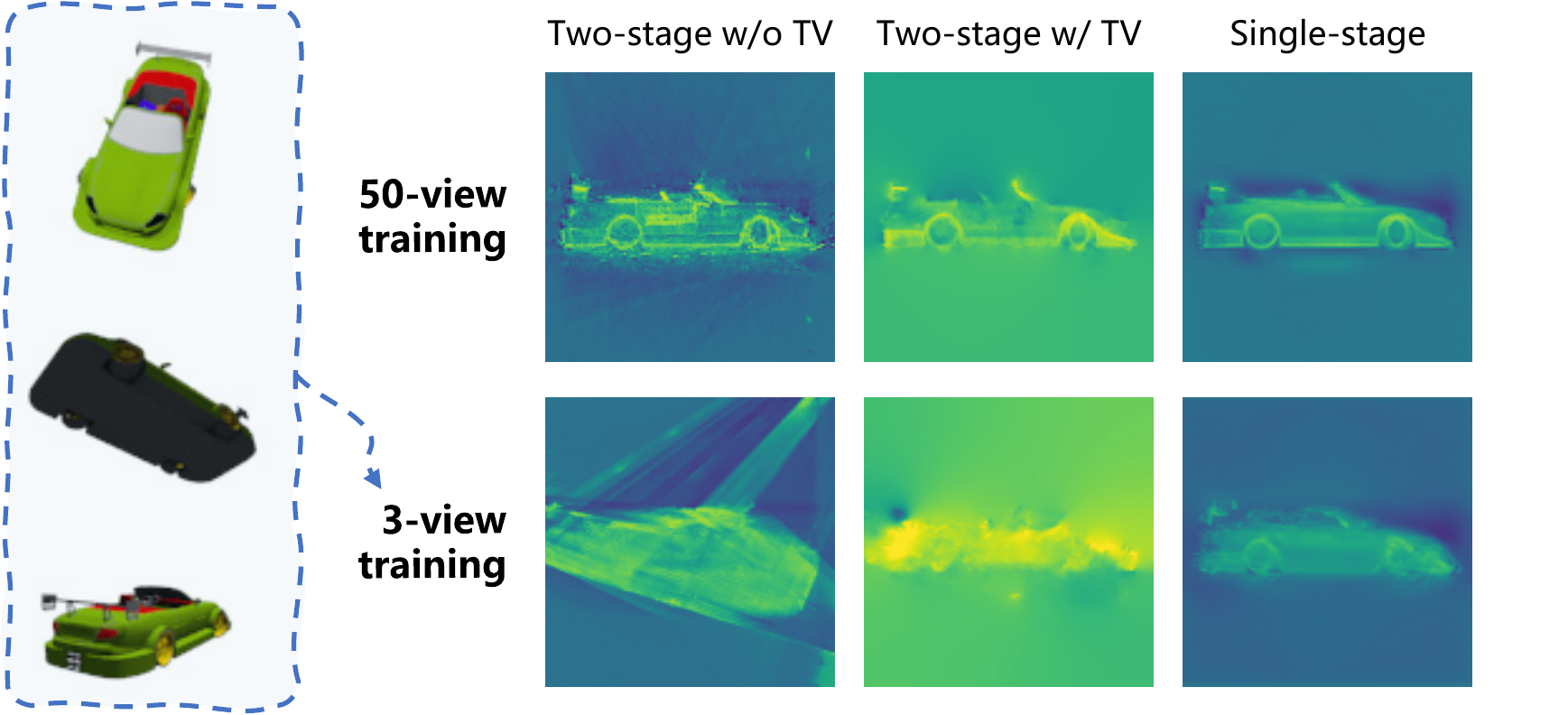}
\end{center}
\caption{Visualization of the scene code $x_\mathrm{XZ}$ at XZ plane. \textbf{Left column:} Two-stage training without TV regularization induces noise and fails in 3-view reconstruction. \textbf{Mid column:} TV regularization imposes smoothing prior at the cost of textural details (top), yet still struggles to cope with sparse views (bottom). \textbf{Right column:} Our single-stage training encourages smooth yet detailed latents and allows for training with sparse views.}
\label{fig:latentviz}
\end{figure}

\paragraph{Limitations of Two-Stage Training for 3D Tasks}
While LDMs with 2D image VAEs are typically trained in two stages~\cite{lsgm,latentdiffusion}, training LDMs with NeRF auto-decoders poses an unprecedented challenge. An expressive latent code is underdetermined when obtained via rendering-based optimization, leading to noisy patterns that distract denoising networks (top-left of Fig.~\ref{fig:latentviz}). Additionally, reconstructing NeRFs from sparse views without a learned prior is exceptionally difficult (bottom-left of Fig.~\ref{fig:latentviz}), limiting training to dense-views settings.

\section{Proposed Method}

\begin{figure*}
\begin{center}
\includegraphics[width=0.87\linewidth]{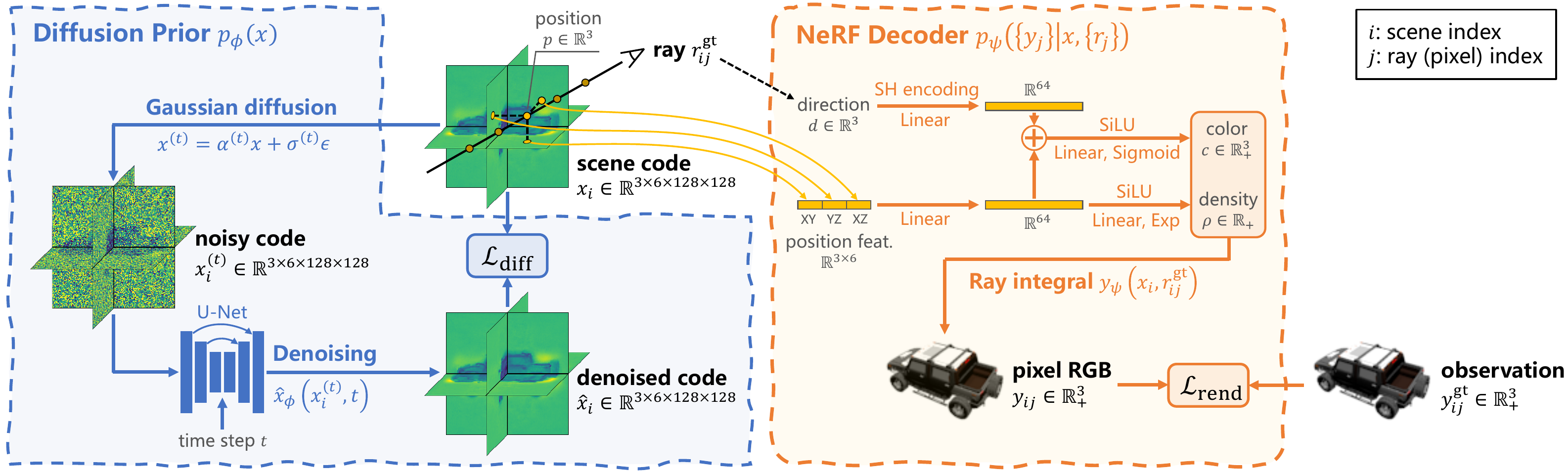}
\end{center}
   \caption{An overview of SSDNeRF framework with a triplane NeRF representation. During training, we feed a batch of observations in the format of RGB values $y^\text{gt}_{ij}$ and rays $r^\text{gt}_{ij}$. The corresponding scene code $x_i$ is randomly initialized and optimized by minimizing the rendering loss $\mathcal{L}_\text{rend}$ and the diffusion loss $\mathcal{L}_\text{diff}$, and model parameters $\phi, \psi$ are also updated along the way.}
\label{fig:framework}
\end{figure*}

To build a holistic model that unifies 3D generation and reconstruction, we propose SSDNeRF, a framework that conjoins the expressive triplane NeRF auto-decoder with a triplane latent diffusion model.
Fig.~\ref{fig:framework} provides an overview of the model. In the following subsections, we elaborate on how training and testing are performed in detail.

\subsection{Single-Stage Diffusion NeRF Training}
\label{sstraining}

An auto-decoder can be regarded as a type of VAE that uses a lookup table encoder instead of the typical neural network encoder. As such, the training objective can be derived in a similar way as for VAEs. 
With NeRF decoder $p_\psi(\{y_j\}|x,\{r_j\})$ and diffusion latent prior $p_\phi(x)$, the training objective is to minimize variational upper bound on the negative log-likelihood (NLL) of observed data $\{y_{ij}^\mathrm{gt}, r_{ij}^\mathrm{gt}\}$~\cite{vae,rezende2014stochastic,lsgm}. In this paper, a simplified training loss is derived by ignoring the uncertainty (variance) in latent codes:
\begin{equation}
    \mathcal{L} = \underbrace{\expect_i{[-\log{p_\psi(\{y_{ij}^\mathrm{gt}\}|x_i,\{r_{ij}^\mathrm{gt}\})}}}_{\text{rendering loss}\ \mathcal{L}_\mathrm{rend}}] + \underbrace{\expect_i{[-\log{p_\phi(x_i)}]}}_{\text{prior term}},
\end{equation}
where the scene codes $\{x_i\}$, prior parameters $\phi$, and decoder parameters $\psi$ are jointly optimized in a single training stage. This loss function consists of the rendering loss $\mathcal{L}_\mathrm{rend}$ in Eq.~(\ref{rendloss}) and a diffusion prior term in the form of NLL. Following \cite{lsgm,diffprior, song2021maximum}, we replace the diffusion NLL with its approximate upper bound $\mathcal{L}_\mathrm{diff}$ in Eq.~(\ref{diffloss}). This technique is also called score distillation in \cite{dreamfusion}. Adding empirical weighting factors, we finalize our training objective as:
\begin{equation}
    \mathcal{L} = \lambda_\mathrm{rend}\mathcal{L}_\mathrm{rend}\left(\{x_i\},\psi\right) + \lambda_\mathrm{diff}\mathcal{L}_\mathrm{diff}\left(\{x_i\},\phi\right).
    \label{mainloss}
\end{equation}

Single-stage training constrains scene codes $\{x_i\}$ with both terms in the loss function, allowing the learned prior to complete the parts unseen to rendering. This is particularly beneficial to training on sparse-view data, where the expressive triplane codes are severely underdetermined.

\paragraph{Balancing Rendering and Prior Weights}
The render-to-prior weight ratio $\lambda_\mathrm{rend} / \lambda_\mathrm{diff}$ is crucial to single-stage training. To make hyperparameters more generalizable to different settings, we design an empirical weighting mechanism, in which the diffusion loss is normalized by the exponential moving average (EMA) of the scene codes' Frobenius norms, expressed as $\lambda_\mathrm{diff} \coloneqq c_\mathrm{diff}/\mathit{EMA}\left(\|x_i\|^2_F\right)$ with a constant scale $c_\mathrm{diff}$, and the rendering weight is determined by the number of views available $N_\mathrm{v}$, expressed as $\lambda_\mathrm{rend} \coloneqq c_\mathrm{rend}(1 - e^{-0.1 N_\mathrm{v}})/N_\mathrm{v}$ with a constant scale $c_\mathrm{rend}$. Intuitively, $N_\mathrm{v}$-based weighting is a calibration to the ray independence assumption in the decoder $p_\psi(\{y_j\}|x,\{r_j\}) \coloneqq \prod_j{\mathcal{N}(y_j|y_\psi(x, r_j), I)} $, preventing the rendering loss from scaling linearly with the number of rays.

\paragraph{Comparison to Two-Stage Generative Neural Fields}
Previous two-stage methods~\cite{gaudi,functa,diffrf,triplanediff} ignore the prior term $\lambda_\mathrm{diff}\mathcal{L}_\mathrm{diff}$ during the first stage of training the auto-decoders. This can be seen as setting the render-to-prior weight ratio $\lambda_\mathrm{rend} / \lambda_\mathrm{diff}$ to infinity, resulting in biased and noisy scene codes ${x_i}$.
Shue~\etal~\cite{triplanediff} partially mitigate this issue by imposing total variation (TV) regularization on triplane scene codes to enforce a smoothing prior, which resembles the LDM constraints on the latent space (mid column of Fig.~\ref{fig:latentviz}). Control3Diff~\cite{control3diff} proposes to learn a conditional diffusion model on data generated by a 3D GAN pretrained on single-view images. 
In contrast, our single-stage training aims to directly incorporate the diffusion prior to promote end-to-end coherence.

\subsection{Image-Guided Sampling and Finetuning}
\label{testmethod}

To achieve generalizable test-time NeRF reconstruction that covers a wide spectrum from single-view to dense observations, we propose performing image-guided sampling and then finetuning the sampled codes considering both the diffusion prior and rendering likelihood.


Following the reconstruction-guided sampling method by Ho~\etal~\cite{ho2022video}, we compute the approximated rendering gradients $g$ \wrt a noisy code $x^{(t)}$, defined as:
\begin{equation}
    g \gets \negthinspace \nabla_{\negthinspace x^{(t)}}\lambda_\mathrm{rend}\smash[b]{\sum_{j}}\frac{1}{2} \left(\negmedspace\frac{\alpha^{(t)}}{\sigma^{(t)}}\negmedspace\right)^{\hspace{-.8ex} 2\omega} \left\|y^\mathrm{gt}_j \negmedspace - y_\psi\left( \negthinspace \hat{x}_\phi(x^{(t)}\negthinspace, t), r^\mathrm{gt}_j \negthinspace \right)\right\|^2 \hspace{-.8ex},
    \label{noisygradients}
\end{equation}
where $\left(\alpha^{(t)} / \sigma^{(t)}\right)^{\hspace{-.1ex}2\omega}$ is an additional weighting factor based on signal-to-noise ratio (SNR), with hyperparameter $\omega$ chosen to be 0.5 or 0.25 in our work. The guidance gradients $g$ are then combined with unconditional score prediction, expressed as a correction to the denoising output $\hat{x}$:
\begin{equation}
    \hat{x} \gets \hat{x} - \lambda_\mathrm{gd} \frac{{\sigma^{(t)}}^2}{\alpha^{(t)}} g
\end{equation}
with guidance scale $\lambda_\mathrm{gd}$. We adopt the predictor-corrector sampler~\cite{song2021scorebased} to solve $x^{(0)}$ by alternating between a DDIM step~\cite{ddim} and multiple Langevin correction steps.

We observe that the reconstruction guidance alone cannot strictly enforce rendering constraints towards faithful reconstruction. To address this issue, we reuse the training objective in Eq.~(\ref{mainloss}) to finetune the sampled scene code $x$, while freezing the diffusion and decoder parameters:
\begin{equation}
\min_{x} \lambda_\mathrm{rend}\mathcal{L}_\mathrm{rend}(x) + \lambda_\mathrm{diff}^\prime\mathcal{L}_\mathrm{diff}(x),
\end{equation}
where $\lambda_\mathrm{diff}^\prime$ is the test-time prior weight, which we find should be lower than the training weight $\lambda_\mathrm{diff}$ for best results, as the prior learned from the training dataset is less reliable when transferred to a different testing dataset. We use Adam~\cite{adam} to optimize the code $x$ for finetuning.

\paragraph{Comparison to Previous NeRF Finetuning Approaches} 
While finetuning with rendering loss is common in view-conditioned NeRF regression methods~\cite{mvsnerf, nerfusion}, our finetuning approach differs in the use of diffusion prior loss on the 3D scene code, which significantly enhances generalization to novel views, as demonstrated in \textsection~\ref{sparseview}. 

\subsection{Implementation Details}


This subsection briefly describes some important technical details. More details can be found in the supplementary.

\paragraph{Prior Gradient Caching}
Triplane NeRF reconstruction requires at least hundreds of optimization iterations on each scene code $x_i$. A problem with the single-stage training loss in Eq.~(\ref{mainloss}) is that the diffusion loss $\mathcal{L}_\mathrm{diff}$ requires much longer time to evaluate than the native NeRF rendering loss $\mathcal{L}_\mathrm{rend}$, reducing overall efficiency. To accelerate reconstruction in both training and test-time finetuning, we introduce a technique called \emph{prior gradient caching}, which caches the back-propagated prior gradients $\nabla_{\negthinspace x} \lambda_\mathrm{diff} \mathcal{L}_\mathrm{diff}$ for re-use in multiple Adam steps, while refreshing the rendering gradients $\nabla_{\negthinspace x} \lambda_\mathrm{rend} \mathcal{L}_\mathrm{rend}$ in each of the steps, which allows for fewer diffusion passes than rendering. A training pseudo-code is given in Algorithm~\ref{algo}.

\paragraph{Denoising Parameterization and Weighting}
The denoising model $\hat{x}_\phi(x^{(t)}, t)$ is implemented as a U-Net~\cite{unet} as in DDPM~\cite{ddpm}, with a total of 122M parameters. Its input and output are noisy and denoised triplane features, respectively, with channels of all three planes stacked together. For the prediction format, we adopt the $v$-parameterization $\hat{v}_\phi(x^{(t)}, t)$ in \cite{velocity}, such that $\hat{x} = \alpha^{(t)}x^{(t)}-\sigma^{(t)}\hat{v}$.
Regarding the weighting function $w^{(t)}$ in the diffusion loss in Eq.~(\ref{diffloss}), LSGM~\cite{lsgm} employs two different mechanisms for optimizing latents ${x_i}$ and diffusion weights $\phi$, respectively, which we find unstable with NeRF auto-decoders. Instead, we observe that the SNR-based weighting $w^{(t)}=\left(\alpha^{(t)} / \sigma^{(t)}\right)^{\hspace{-.1ex}2\omega}$ used in Eq.~(\ref{noisygradients}) works well with our models.

\section{Experiments}

\subsection{Datasets}

We conduct experiments on the ShapeNet SRN~\cite{shapenet2015, srn} and Amazon Berkeley Objects (ABO) Tables~\cite{abo} datasets for benchmarking with previous work.
The SRN dataset provides single-object scenes in two categories, \ie, Cars and Chairs, with a train/test split of 2458/703 for Cars and 4612/1317 for Chairs. Each train scene has 50 random views from a sphere and each test scene has 251 spiral views from the upper hemisphere. The ABO Tables dataset
provides a train/test split of 1520/156 table scenes, where each scene has 91 views from the upper hemisphere. For both datasets, we use the provided renderings (resized to 128\texttimes128) with ground truth poses for training and testing.

\begin{algorithm}[t]
\DontPrintSemicolon
\KwIn{$\{y_{ij}^\mathrm{gt}, r_{ij}^\mathrm{gt}\}$}
Initialize $\{x_i\}, \phi, \psi$ \\
\For(\tcp*[f]{outer loop of $K_\mathrm{out}$ iterations}){$k_\mathrm{out} \coloneqq 1\cdots K_\mathrm{out}$}{
    Sample a batch of scenes $i \in B_\mathrm{sc}$ \\
    $g_\phi, g_x^\mathrm{diff} \gets \nabla_{\negthinspace \phi, \{x_i\}_{B_\mathrm{sc}}} \lambda_\mathrm{diff} \mathcal{L}_\mathrm{diff}$ \tcp*{diffusion grad}
    $\phi \gets \phi - \mathit{Adam}(g_\phi)$ \\
    \For(\tcp*[f]{inner loop of $K_\mathrm{in}$ iterations}){$k_\mathrm{in} \coloneqq 1\cdots K_\mathrm{in}$}{ 
        Sample a batch of rays $j \in B_\mathrm{ray}$ \\
        $g_x^\mathrm{rend} \gets \nabla_{\negthinspace \{x_i\}_{B_\mathrm{sc}}} \lambda_\mathrm{rend} \mathcal{L}_\mathrm{rend}$ \tcp*{rendering grad}
        $g_x \gets g_x^\mathrm{rend} + g_x^\mathrm{diff}$ \tcp*{add cached prior grad}
        $\{x_i\}_{B_\mathrm{sc}} \gets \{x_i\}_{B_\mathrm{sc}} - \mathit{Adam}(g_x)$ \\
        \If(\tcp*[f]{last inner iteration}){$k_\mathrm{in} = K_\mathrm{in}$}{
            $g_\psi \gets \nabla_{\negthinspace \psi} \lambda_\mathrm{rend} \mathcal{L}_\mathrm{rend}$ \\
            $\psi \gets \psi - \mathit{Adam}(g_\psi)$
            }
        }
    }
    
\caption{Single-stage diffusion NeRF training}
\label{algo}
\end{algorithm}

\begin{figure*}[t]
\begin{center}
\includegraphics[width=1.0\linewidth]{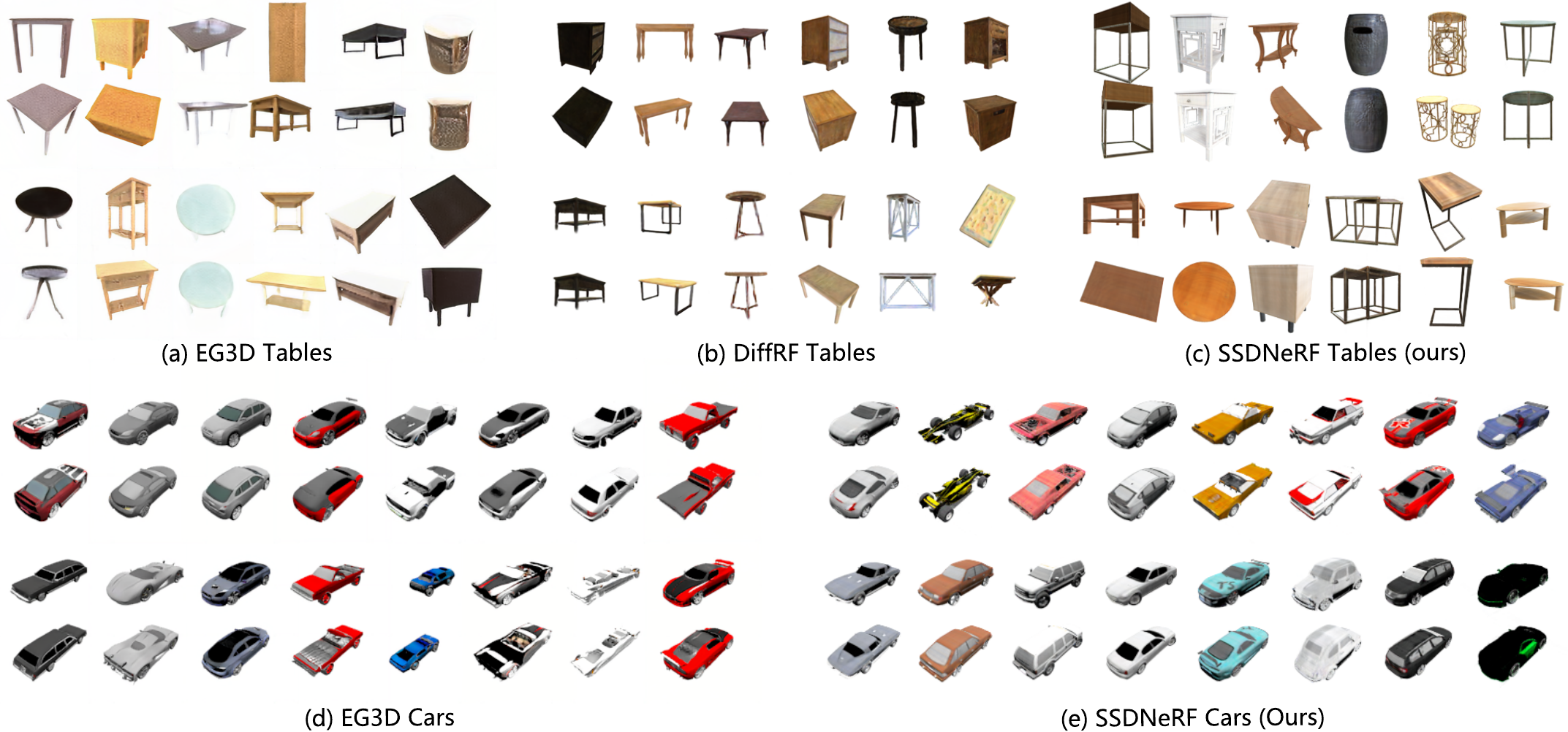}
\end{center}
\caption{Qualitative comparison between unconditional generative models trained on ABO Tables and SRN Cars.}
\label{fig:uncond}
\end{figure*}

\subsection{Unconditional Generation}

\addtocounter{footnote}{-1}

\begin{table}[t]
\begin{center}
\scalebox{0.85}{%
\setlength{\tabcolsep}{0.3em}
\begin{tabular}{lccccc}
\toprule
\multirow{2}[2]{*}{Method} & \multirow{2}[2]{*}{Type} & \multicolumn{2}{c}{Cars} & \multicolumn{2}{c}{Tables} \\
\cmidrule(lr){3-4}
\cmidrule(lr){5-6}
& & FID\makebox[0pt][l]{\textdownarrow}\pmplaceholder & KID{\scriptsize/10\textsuperscript{\textminus3}}\textdownarrow & FID\makebox[0pt][l]{\textdownarrow}\pmplaceholder & KID{\scriptsize/10\textsuperscript{\textminus3}}\textdownarrow \\
\midrule
Functa~\cite{functa} & LDM & 80.3\phantom{0}\pmplaceholder & - & - & - \\
$\pi$-GAN~\cite{piGAN} & GAN & 36.7\ndagger\phantom{0}\pmplaceholder & - & 41.67\nsection\pmplaceholder & 13.82\nsection\pmplaceholder \\
EG3D~\cite{eg3d} & GAN & \textbf{10.46}\nast\pmplaceholder & 4.90\nast\pmplaceholder & 31.18\nsection\pmplaceholder & 11.67\nsection\pmplaceholder \\
DiffRF~\cite{diffrf} & LDM & - & - & 27.06\pmplaceholder & 10.03\pmplaceholder \\
\midrule
Ours (2-stage) & LDM & 16.33{\scriptsize\textpm0.93} & 6.38{\scriptsize\textpm0.41} & - & - \\
Ours (1-stage) & LDM & 11.08{\scriptsize\textpm1.11} & \textbf{3.47}{\scriptsize\textpm0.23} & \textbf{14.27}{\scriptsize\textpm0.66} & \textbf{\phantom{0}4.08}{\scriptsize\textpm0.33} \\ 
\bottomrule
\end{tabular}}
\end{center}
\caption{Unconditional generation results on SRN Cars and ABO Tables. \textdagger\ denotes results reported by Functa~\cite{functa}. \textsection\ denotes results reported by DiffRF~\cite{diffrf}. \textasteriskcentered\ denotes results reproduced by us using the official public code with a bugfix.\negmedspace\protect\footnotemark\ We show \textpm2$\sigma$ intervals.
}
\label{tab:uncond}
\end{table}

\addtocounter{footnote}{-1}
\stepcounter{footnote}\footnotetext{\url{https://github.com/nvlabs/eg3d/issues/67}}

\begin{table*}[t]
\begin{center}
\scalebox{0.82}{%
\setlength{\tabcolsep}{0.55em}
\begin{tabular}{lcccccccccccccccc}
\toprule
\multirow{2}[2]{*}{Method} & \multicolumn{4}{c}{Cars 1-view} & \multicolumn{4}{c}{Cars 2-view} & \multicolumn{4}{c}{Chairs 1-view} & \multicolumn{4}{c}{Chairs 2-view}\\
\cmidrule(lr){2-5}
\cmidrule(lr){6-9}
\cmidrule(lr){10-13}
\cmidrule(lr){14-17}
& \makebox[0pt]{PSNR\textuparrow} & \makebox[0pt]{SSIM\textuparrow} & \makebox[0pt]{LPIPS\textdownarrow} & \makebox[0pt]{FID\textdownarrow} & \makebox[0pt]{PSNR\textuparrow} & \makebox[0pt]{SSIM\textuparrow} & \makebox[0pt]{LPIPS\textdownarrow} & \makebox[0pt]{FID\textdownarrow} & \makebox[0pt]{PSNR\textuparrow} & \makebox[0pt]{SSIM\textuparrow} & \makebox[0pt]{LPIPS\textdownarrow} & \makebox[0pt]{FID\textdownarrow} & \makebox[0pt]{PSNR\textuparrow} & \makebox[0pt]{SSIM\textuparrow} & \makebox[0pt]{LPIPS\textdownarrow} & \makebox[0pt]{FID\textdownarrow} \\
\midrule
3DiM~\cite{3dim} & 21.01 & 0.57 & - & \phantom{0}\textbf{8.99} & - & - & - & - & 17.05 & 0.53 & - & \phantom{0}\textbf{6.57} & - & - & - & - \\
PixelNeRF~\cite{pixelnerf} & 23.17 & 0.90 &  0.146\nddagger & 59.24\ndagger & 25.66 & \textbf{0.94} & - & - & 23.72 & 0.91 & 0.128\nddagger & 38.49\ndagger & 26.20 & 0.94 & - & - \\
SRN~\cite{srn} & 22.25\nsection & 0.89\nsection & 0.129\nddagger & 41.21\ndagger & 24.84\nsection & 0.92\nsection & - & - & 22.89\nsection & 0.89\nsection & 0.104\nddagger & 26.51\ndagger & 24.48\nsection & 0.92\nsection & - & -\\
CodeNeRF~\cite{codenerf} & \textbf{23.80} & \textbf{0.91} & 0.118\nast & 56.34\nast & 25.71 & 0.93 & 0.108\nast & 56.13\nast & 23.66 & 0.90 & 0.106\nast & 31.65\nast & 25.63 & 0.91 & 0.097\nast & 29.90\nast \\
VisionNeRF~\cite{visionnerf} & 22.88 & \textbf{0.91} & 0.084 & 21.31\ndagger & - & - & - & - & \textbf{24.48} & \textbf{0.93} & 0.077 & 10.05\ndagger & - & - & - & - \\
\midrule
Ours (1-stage) & 23.52 & \textbf{0.91} & \textbf{0.078} & 16.39 & \textbf{26.49} & \textbf{0.94} & \textbf{0.054} & \textbf{10.66} & 24.35 & \textbf{0.93} & \textbf{0.067} & 10.13 & \textbf{26.94} & \textbf{0.95} & \textbf{0.055} & \textbf{10.85} \\
\bottomrule
\end{tabular}}
\end{center}
\caption{Single-view and two-view reconstruction results on SRN Cars and Chairs. For consistency with prior work, we use view \#64 of the test scene as single-view input and view \#64 and \#104 as two-view input. \textdagger\ denotes results reported by 3DiM~\cite{3dim}. \textdaggerdbl\ denotes results reported by VisionNeRF~\cite{visionnerf}, \textsection\ denotes results reported by PixelNeRF~\cite{codenerf}, \textasteriskcentered\ denotes results reproduced by us using the official code. -~indicates results are unavailable.} 
\label{tab:recsota}
\end{table*}

In this section, we conduct evaluations for unconditional generation using the SRN Cars and ABO Tables dataset. The Cars dataset poses a challenge in generating sharp and intricate textures, whereas the Tables dataset comprises of diverse geometries with realistic materials. Models are trained on all images of the training set for 1M iterations.

\paragraph{Evaluation Protocol and Metrics}
For SRN Cars, following Functa~\cite{functa}, we sample 704 scenes from the diffusion model, and render each scene using the fixed 251 camera poses from the test set. For ABO Tables, following DiffRF~\cite{diffrf}, we sample 1000 scenes and render each scene with 10 random cameras. We adopt standard generation metrics including Fréchet Inception Distance (FID)~\cite{FID} and Kernel Inception Distance (KID)~\cite{KID}. The metrics' reference sets are all images in the test set for SRN Cars and all images in the entire dataset for ABO Tables, respectively.

\paragraph{Comparison to the State of the Art}
As shown in Table~\ref{tab:uncond}, on SRN Cars, SSDNeRF (1-stage) outperforms EG3D in KID (a more suitable measure for small datasets) by a clear margin. Meanwhile, its FID is drastically better than Functa, which uses an LDM but with low dimensional latent codes. On ABO Tables, SSDNeRF shows significantly better performance than EG3D and DiffRF.

\paragraph{Single- \vs Two-stage}
On SRN Cars, we compare the proposed single-stage training against two-stage training with tuned TV regularization using the same model architecture. The results in Table~\ref{tab:uncond} indicate substantial advantage of single-stage training (KID{\scriptsize/10\textsuperscript{\textminus3}} 3.47 \vs 6.38). 

\paragraph{Qualitative Results}
As shown in Fig.~\ref{fig:uncond}, SSDNeRF generates more regular geometries than the slightly skewed and distorted shapes by EG3D~\cite{eg3d}. Compared to DiffRF~\cite{diffrf}, our method produces sharp details and reflective materials, thanks to our more expressive model with latents of higher spatial resolution and view-dependent NeRF decoder. 

\subsection{Sparse-View NeRF Reconstruction}
\label{sparseview}

This section presents experiments on 3D reconstruction from sparse-view images of unseen objects in SRN Cars and Chairs test sets. The Cars dataset presents the challenge of recovering distinct textures, while the Chairs dataset requires accurate reconstruction of diverse shapes. Models are trained on all images of the training set for 80K iterations, as we find that longer schedule leads to decaying performance in reconstructing unseen objects. This behaviour is in accordance with the interpolation results in \textsection~\ref{interpolation}.

\paragraph{Evaluation Protocol and Metrics}
We use the evaluation protocol and metrics in PixelNeRF~\cite{pixelnerf}. Given input images sampled from each test scene, we obtain the triplane scene code via guidance-finetuning and evaluate novel view synthesis quality with respect to the unseen images. The image quality metrics include average peak signal-to-noise-ratio (PSNR), structural similarity (SSIM)~\cite{ssim}, and Learned Perceptual Image Patch Similarity (LPIPS)~\cite{lpips}. In addition, we evaluate the FID between all synthesized images and ground truth images as in 3DiM~\cite{3dim}.

\paragraph{Comparison to the State of the Art}

\begin{figure*}[t]
\begin{center}
\includegraphics[width=1.0\linewidth]{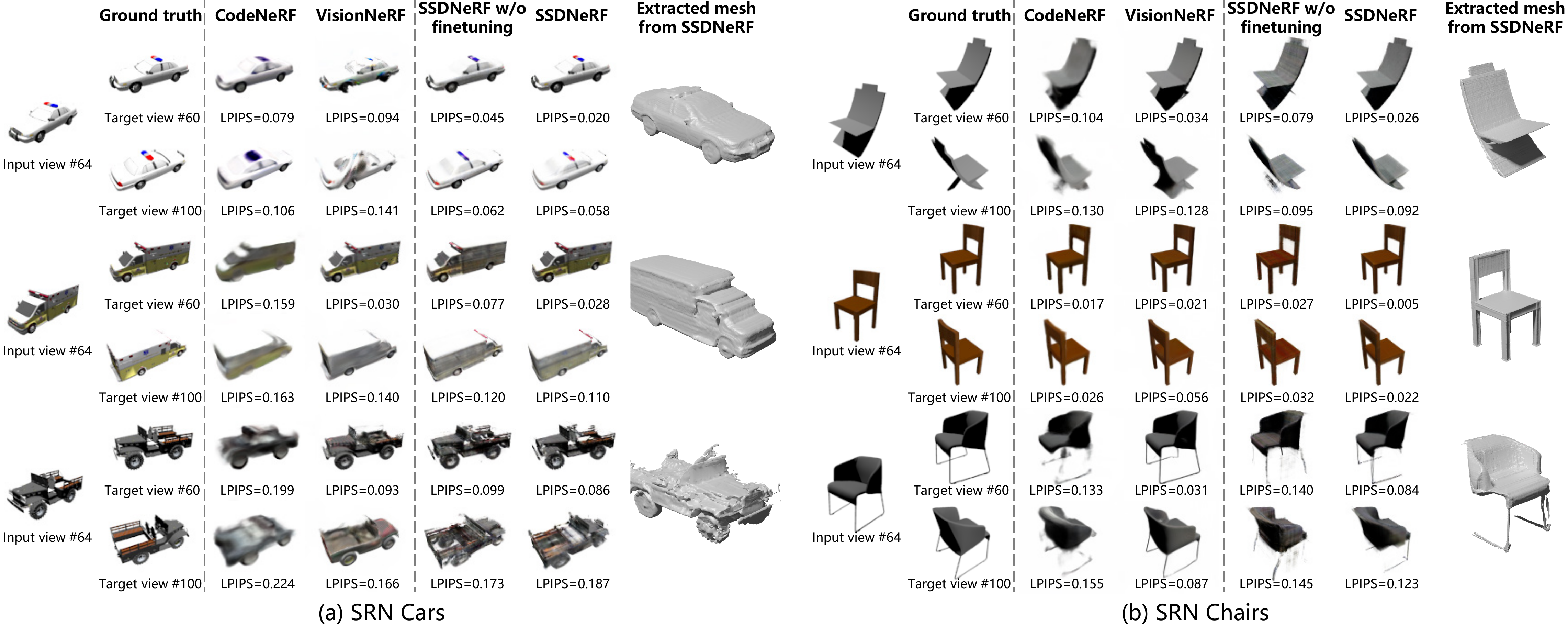}
\end{center}
\vspace{-0.3ex}
\caption{Qualitative comparison of single-view reconstruction methods on unseen test objects in SRN Cars and Chairs.}
\label{fig:singleview}
\end{figure*}

Table~\ref{tab:recsota} compares SSDNeRF against previous approaches in single-view and two-view reconstruction settings. Overall, SSDNeRF reaches the best LPIPS of all tasks, indicating the best perceptual fidelity. In contrast, 3DiM generates high quality images (best FID) but with the lowest fidelity to the ground truth (lowest PSNR); CodeNeRF reports the best PSNR on single-view Cars, but its limited expressiveness leads to blurry outputs (Fig.~\ref{fig:singleview}) and less competitive LPIPS and FID; VisionNeRF achieves a balanced performance on all single-view metrics, but may struggle to generate textural details on the unseen side of cars (\eg, the other side of the ambulance in Fig.~\ref{fig:singleview}). Moreover, SSDNeRF exhibits a clear advantage in two-view reconstruction, achieving the best performance on all relevant metrics.

\paragraph{Single- \vs Two-stage}

As demonstrated in Table~\ref{tab:recablation}, the model trained in a single stage (A0) outperforms the same architecture trained in two stages with TV regularization (A1) in all metrics of single-view reconstruction.

\paragraph{Ablation Studies on Test-Time Finetuning}

As shown in Table~\ref{tab:recablation}, we evaluate the effectiveness of test-time finetuning and the contribution of the learned diffusion prior with two ablation experiments: (A2) removing the diffusion loss during finetuning and using only the rendering loss, and (A3) omitting the finetuning process entirely. The results indicate that finetuning with single-view rendering loss provides only marginal improvements over guided sampling (A2 \vs A3), while the learned diffusion prior significantly boosts the LPIPS and FID scores (A0 \vs A2), highlighting its importance in recovering sharp and distinct contents. Moreover, the qualitative results in Fig.~\ref{fig:singleview} reveal that views with higher overlap to the input view benefit the most from finetuning, meeting our expectation that finetuning helps faithfully reconstruct the exact observations.

\begin{table}[t]
\vspace{-1ex}
\begin{center}
\setlength{\tabcolsep}{0.3em}
\scalebox{0.85}{%
\begin{tabular}{cllcccc}
\toprule
ID & Training & Finetuning & PSNR\textuparrow & SSIM\textuparrow & LPIPS\textdownarrow & FID\textdownarrow \\
\midrule
A0 & 1-stage & Rend + Diff & 23.52 & 0.913 & 0.078 & 16.39 \\
A1 & 2-stage & Rend + Diff & 22.83 & 0.906 & 0.090 & 20.97 \\
A2 & 1-stage & Rend & 23.13 & 0.907 & 0.088 & 27.93 \\
A3 & 1-stage & None & 23.07 & 0.905 & 0.092 & 30.95 \\
\bottomrule
\end{tabular}}
\end{center}
\caption{Ablation results on single-view reconstruction of SRN Cars.}
\label{tab:recablation}
\end{table}

\begin{figure}[t]
\begin{center}
\vspace{-1ex}
\includegraphics[width=0.75\linewidth]{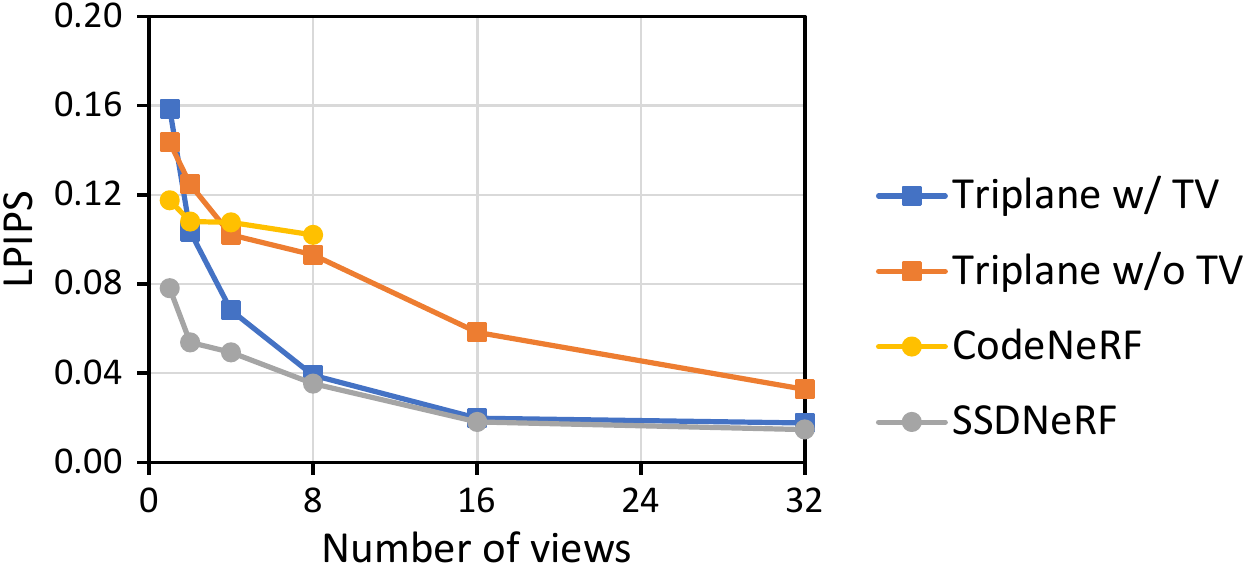}
\end{center}
\caption{LPIPS scores (lower is better) of novel view synthesis from sparse-to-dense inputs, evaluated on SRN Cars test set. The triplane baselines adopt mean initialization for better performance.}
\label{fig:sparse2dense}
\end{figure}

\paragraph{Sparse-to-Dense Reconstruction}

To validate that SSDNeRF seamlessly bridges sparse- and dense-view NeRF reconstruction, we evaluate its novel view synthesis performance with the number of input views varying from 1 to 32. We compare our model to the triplane NeRF baseline trained as an auto-decoder with optional TV regularization instead of diffusion prior. Meanwhile, we also evaluate CodeNeRF~\cite{codenerf}, an auto-decoder with 256-d latent codes. The results in Fig.~\ref{fig:sparse2dense} show that SSDNeRF excels in all settings, especially in 1 to 4 views.
In contrast, CodeNeRF is outperformed by vanilla triplane NeRF with more views.  

\subsection{Training SSDNeRF on Sparse-View Dataset}

In this section, we train SSDNeRF on a sparse-view subset of the full SRN Cars training set, in which a fixed set of only three views are randomly picked from each scene. Note that a reasonable decline in performance compared to dense-view training is expected as the whole training dataset has been reduced to 6\% of its original size.

\begin{figure}[t]
\begin{center}
\includegraphics[trim={0 0 0 2.1cm}, clip, width=1.0\linewidth]{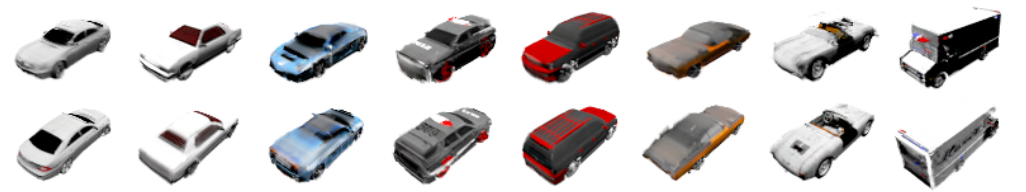}
\end{center}
\caption{Images generated by SSDNeRF trained on a 3-view subset of SRN Cars training set.}
\label{fig:uncond_sparse}
\end{figure}

\paragraph{Unconditional Generation}
We adopt a training trick that resets the triplane codes to their mean value halfway through training. This helps to prevent the model from getting stuck in a local minimum that overfits geometric artifacts. We also double the length of the training schedule accordingly. 
The model achieves a decent FID of 19.04{\scriptsize\textpm1.10} and a KID{\scriptsize/10\textsuperscript{\textminus3}} of 8.28{\scriptsize\textpm0.60}. Results are visualized in Fig.~\ref{fig:uncond_sparse}.

\paragraph{Single-View Reconstruction}
We adopt the same training strategy as in \textsection~\ref{sparseview}. With our guidance-finetuning approach, the model achieves an LPIPS score of 0.106, even outperforming most of the previous methods in Table~\ref{tab:recsota} that use the full training set.

\begin{figure}[t]
\begin{center}
\includegraphics[width=1.0\linewidth]{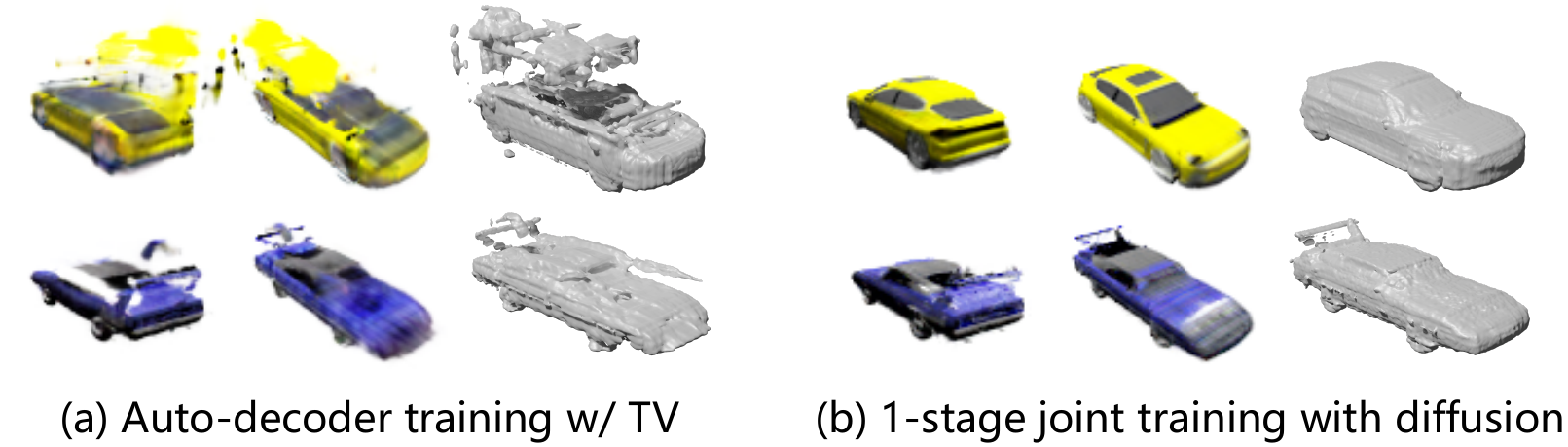}
\end{center}
   \caption{Qualitative comparison between scene codes learned from 3 views by (a) triplane auto-decoder with TV regularization \vs (b) single-stage diffusion NeRF.}
\label{fig:3viewtrain}
\end{figure}

\paragraph{Comparison to TV Regularization}
Fig.~\ref{fig:3viewtrain}~(b) shows the RGB images and geometries represented by the scene latent codes learned from three views during training. By comparison, vanilla triplane auto-decoder with TV regularization (Fig.~\ref{fig:3viewtrain}~(a)) often fails to reconstruct a scene from sparse views, leading to severe geometric artifacts. As a result, previously it has been infeasible to train two-stage models with expressive latents on sparse-view data.

\subsection{NeRF Interpolation}
\label{interpolation}

\begin{figure}[t]
\begin{center}
\includegraphics[width=1.0\linewidth]{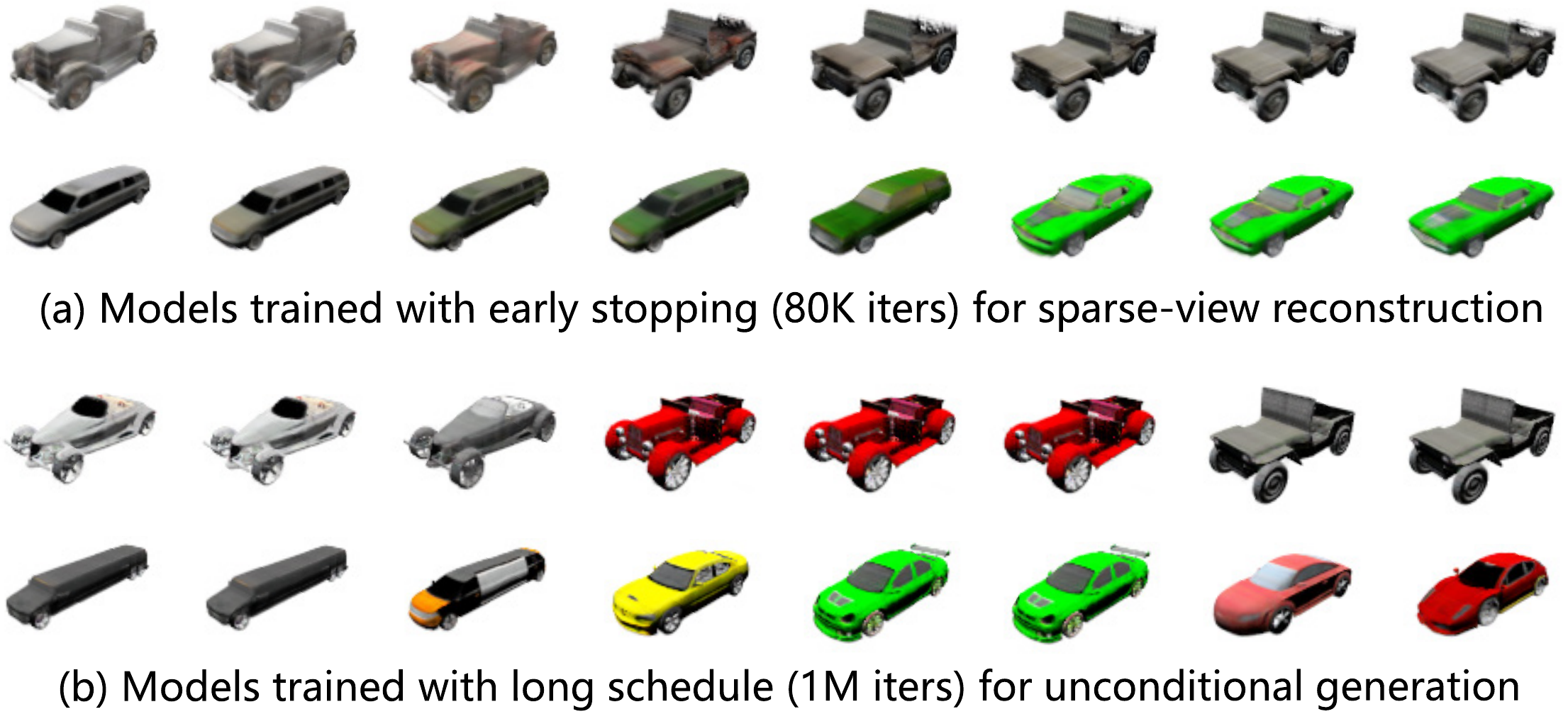}
\end{center}
\caption{Interpolation between the leftmost and rightmost samples using DDIM~\cite{ddim}.}
\label{fig:interp}
\end{figure}

Following DDIM~\cite{ddim}, we can sample two initial values $x^{(T)}\sim\mathcal{N}(0, I)$, interpolate them using spherical linear interpolation~\cite{shoemake1985animating}, and then use the deterministic solver to obtain interpolated samples. However, as noted by \cite{preechakul2021diffusion, rissanen2022generative}, standard Gaussian diffusion models often result in non-smooth interpolation. In SSDNeRF (with results shown in Fig.~\ref{fig:interp}), we find that the model (a) trained with early stopping for sparse-view reconstruction produces reasonably smooth transitions between samples, while the model (b) trained with a longer schedule for unconditional generation produces distinct yet discontinuous samples. This suggests that early stopping preserves a smoother prior, leading to better generalization for sparse-view reconstruction.

\section{Conclusion}

In this paper, we propose SSDNeRF, which combines the diffusion model and NeRF representation through a novel single-stage training paradigm with an end-to-end justifiable loss. Notably, it overcomes the limitations in previous work where implicit neural fields must be obtained from dense observations first, before training the diffusion models to learn their manifold. With strong performance on multiple benchmarks, SSDNeRF demonstrates a significant advancement towards a unified framework for general 3D content manipulation.

\paragraph{Limitations and Future Work}
Currently, our method relies on ground truth camera parameters during both training and testing. Future work may explore transform-invariant models. Additionally, the diffusion prior can become discontinuous with prolonged training, which affects generalization. 
Although early stopping is temporarily used, a better network design
or a larger training dataset
may be able to address this problem fundamentally. 

\ificcvfinal

\paragraph{Acknowledgements}

We thank Norman Müller for sharing the baseline results on ABO Tables. Hansheng Chen and Wei Tian acknowledge the funding by the National Natural Science Foundation of China (No.~52002285), the Shanghai Science and Technology Commission (No.~21ZR1467400), the original research project of Tongji University (No.~22120220593), the National Key R\&D Program of China (No.~2021YFB2501104), and the Natural Science Foundation of Chongqing (No.~2023NSCQ-MSX4511).

\fi

\bib

%% file: sections/supp_sections.tex
\setcounter{section}{0}
\renewcommand{\thesection}{\Alph{section}}

\section{Details on Batch-Wise Rendering Loss}

During single-stage training and test-time reconstruction, we randomly sample a batch of rays $B_\mathrm{ray}$ from all available observations for each rendering pass. The actual rendering loss needs to be rescaled to account for the batch size $|B_\mathrm{ray}|$.

For single-stage training and test-time \emph{finetuning} based on Adam~\cite{adam}, we rescale the rendering loss to keep its overall magnitude invariant to the batch size $|B_\mathrm{ray}|$:
\begin{equation}
    \mathcal{L}_\mathrm{rend}(\{x_i\},\psi) = \expect_{i}\left[\frac{{N_\mathrm{ray}}_i}{|B_\mathrm{ray}|} \smash[b]{\sum_{j \in B_\mathrm{ray}}} \negthickspace {\frac{1}{2}\left\|y^\mathrm{gt}_{ij} - y_\psi\left(x_i, r^\mathrm{gt}_{ij}\right)\right\|^2}\right],
    \label{actualrendloss}
\end{equation}
where ${N_\mathrm{ray}}_i$ is the total number of observed rays of the $i$-th scene.

For test-time gradient \emph{guidance}, however, we treat the sampled batch $B_\mathrm{ray}$ as if it constitutes the full observation set. Thus, the gradients originally defined in Eq.~(5) are actually calculated by:
\begin{equation}
    g \gets \negthinspace \nabla_{\negthinspace x^{(t)}}\lambda^{B_\mathrm{ray}}_\mathrm{rend} \hspace{-.8ex} \smash[b]{\sum_{j \in B_\mathrm{ray}}} \hspace{-.9ex} \frac{1}{2} \left(\negmedspace\frac{\alpha^{(t)}}{\sigma^{(t)}}\negmedspace\right)^{\hspace{-.8ex} 2\omega} \left\|y^\mathrm{gt}_j \negmedspace - y_\psi\left(\negthinspace\hat{x}_\phi(x^{(t)}\negthinspace, t), r^\mathrm{gt}_j\negthinspace\right)\right\|^2 \hspace{-.8ex},
    \label{actualnoisygradients}
\end{equation}
in which the balanced rendering weight $\lambda^{B_\mathrm{ray}}_\mathrm{rend} \coloneqq c_\mathrm{rend}(1 - e^{-0.1 N^{B_\mathrm{ray}}_\mathrm{v}})/N^{B_\mathrm{ray}}_\mathrm{v}$
is determined by the batch-effective number of views $N^{B_\mathrm{ray}}_\mathrm{v}$ instead of the number of all available views $N_\mathrm{v}$, with their relationship defined as:
\begin{equation}
N^{B_\mathrm{ray}}_\mathrm{v} = \frac{|B_\mathrm{ray}|}{N_\mathrm{ray}} N_\mathrm{v},
\end{equation}
where $N_\mathrm{ray}$ is the total number of observed rays of a test scene.

\section{Implementation and Hyperparameters}

\begin{table*}[t]
\begin{center}
\scalebox{0.85}{%
\setlength{\tabcolsep}{0.19em}
\begin{tabular}{lcccccc}
\toprule
& \multicolumn{3}{c}{Unconditional} & \multicolumn{3}{c}{Reconstruction} \\
\cmidrule(lr){2-4} \cmidrule(lr){5-7}
& Cars (full) & Cars (3-view) & Tables (full) & Cars (full) & Cars (3-view) & Chairs (full) \\
\midrule
$x$ shape & \multicolumn{6}{c}{\hspace{1cm} 3\texttimes6\texttimes128\texttimes128} \\
Latent dimensionality $\dim{(X)}$ & \multicolumn{6}{c}{\hspace{1cm} 294912} \\
U-Net base channels & \multicolumn{6}{c}{\hspace{1cm} 128} \\
U-Net channel multiplier & \multicolumn{6}{c}{\hspace{1cm} 1, 2, 2, 4, 4} \\
U-Net depth & \multicolumn{6}{c}{\hspace{1cm} 2} \\
U-Net attention resolutions & \multicolumn{6}{c}{\hspace{1cm} 32, 16, 8} \\
U-Net attention heads & \multicolumn{6}{c}{\hspace{1cm} 4} \\
U-Net dropout & 0.0 & 0.0 & 0.0 & 0.1 & 0.1 & 0.1 \\ 
Diffusion steps & \multicolumn{6}{c}{\hspace{1cm} 1000} \\
Noise schedule & \multicolumn{6}{c}{\hspace{1cm} Linear} \\
\midrule
Scene batch size $|B_\mathrm{sc}|$ & \multicolumn{6}{c}{\hspace{1cm} 16} \\
Ray batch size $|B_\mathrm{ray}|$ & \multicolumn{6}{c}{\hspace{1cm} 4096} \\
Rendering weight constant $c_\mathrm{rend}$ & \multicolumn{6}{c}{\hspace{1cm} 40 \texttimes\ 2\textsuperscript{\textminus14}} \\
Diffusion weight constant $c_\mathrm{diff}$ & \multicolumn{6}{c}{\hspace{1cm} $4$} \\
SNR power $\omega$ & 0.5 & 0.5 & 0.5 & 0.5 & 0.5 & 0.25 \\
Outer loop iterations $K_\mathrm{out}$ & 1M & 2M & 1M & 80K & 80K & 80K \\
Inner loop iterations $K_\mathrm{in}$ & \scalebox{0.72}{$
\begin{dcases} 
16, & k_\mathrm{out} \leq \text{2K}, \\ 
4, & \text{2K} < k_\mathrm{out} \leq \text{100K}, \\
2, & k_\mathrm{out} > \text{500K}.
\end{dcases}
$} & \scalebox{0.72}{$
\begin{dcases} 
16, & k_\mathrm{out} \leq \text{2K}, \\ 
2, & k_\mathrm{out} > \text{2K}.
\end{dcases}
$} & \scalebox{0.72}{$
\begin{dcases} 
16, & k_\mathrm{out} \leq \text{2K}, \\ 
4, & \text{2K} < k_\mathrm{out} \leq \text{100K}, \\
2, & k_\mathrm{out} > \text{500K}.
\end{dcases}
$} & \scalebox{0.72}{$
\begin{dcases} 
16, & k_\mathrm{out} \leq \text{2K}, \\ 
4, & k_\mathrm{out} > \text{2K}.
\end{dcases}
$} & \scalebox{0.72}{$
\begin{dcases} 
16, & k_\mathrm{out} \leq \text{2K}, \\ 
2, & k_\mathrm{out} > \text{2K}.
\end{dcases}
$} & \scalebox{0.72}{$
\begin{dcases} 
16, & k_\mathrm{out} \leq \text{2K}, \\ 
4, & k_\mathrm{out} > \text{2K}.
\end{dcases}
$} \\
Latent base learning rate & 0.005\phantom{0} & 0.005\phantom{0} & 0.003\phantom{00} & 0.01\phantom{00} & 0.01\phantom{00} & 0.01\phantom{00} \\
Decoder base learning rate & 0.001\phantom{0} & 0.001\phantom{0} & 0.0006\phantom{0} & 0.001\phantom{0} & 0.001\phantom{0} & 0.001\phantom{0} \\
Diffusion base learning rate & 0.0001 & 0.0001 & 0.00006 & 0.0001 & 0.0001 & 0.0001 \\
Learning rate multiplier & \scalebox{0.72}{$
\begin{dcases} 
1, & k_\mathrm{out} \leq \text{500K}, \\ 
0.5, & k_\mathrm{out} > \text{500K}.
\end{dcases}
$} & \scalebox{0.72}{$
\begin{dcases} 
1, & k_\mathrm{out} \leq \text{500K}, \\ 
0.5, & \text{500K} < k_\mathrm{out} \leq \text{1M}, \\
1, & \text{1M} < k_\mathrm{out} \leq \text{1.5M}, \\
0.5, & k_\mathrm{out} > \text{1.5M}.
\end{dcases}
$} & \scalebox{0.72}{$
\begin{dcases} 
1, & k_\mathrm{out} \leq \text{500K}, \\ 
0.5, & k_\mathrm{out} > \text{500K}.
\end{dcases}
$} & 1 & 1 & 1 \\
\midrule
Ray batch size $|B_\mathrm{ray}|$ & \multicolumn{6}{c}{\hspace{1cm} 16384} \\
DDIM steps & 50 & 50 & 50 & 75 & 75 & 75 \\
Langevin inner iterations & 0 & 0 & 0 & 0 & 0 & 5 \\
Langevin step size $\delta$ & \multicolumn{6}{c}{\hspace{1cm} 0.4} \\
Guidance scale $\lambda_\mathrm{gd}$ & - & - & - & 3.2 \texttimes\ 2\textsuperscript{14} & 0.8 \texttimes\ 2\textsuperscript{14} & 0.4 \texttimes\ 2\textsuperscript{14} \\
Rendering weight constant $c_\mathrm{rend}$ & \multicolumn{6}{c}{\hspace{1cm} 40 \texttimes\ 2\textsuperscript{\textminus14}} \\
FT Diffusion weight constant $c^\prime_\mathrm{diff}$ & \multicolumn{6}{c}{\hspace{1cm} $1$}\\
FT SNR power $\omega$ & 0.5 & 0.5 & 0.5 & 0.5 & 0.5 & 0.25 \\
FT outer loop iterations $K_\mathrm{out}$ & 0 & 0 & 0 & Table~\ref{tab:params2} & Table~\ref{tab:params2} & Table~\ref{tab:params2}\\
FT inner loop iterations $K_\mathrm{in}$ & \multicolumn{6}{c}{\hspace{1cm} Table~\ref{tab:params2}} \\
FT latent base learning rate & \multicolumn{6}{c}{\hspace{1cm} Table~\ref{tab:params2}} \\
FT learning rate multiplier & \multicolumn{6}{c}{\hspace{1cm} $0.998^{k_\mathrm{out} \cdot K_\mathrm{in} + k_\mathrm{in}}$} \\
\bottomrule
\end{tabular}}
\end{center}
\caption{Architecture/training/testing hyperparameters. $k_\mathrm{out}, k_\mathrm{in}$ correspond to the outer and inner loop iteration indices in Algorithm 1. 2\textsuperscript{14} is the number of pixels per view.}
\label{tab:params}
\end{table*}

\subsection{Implementation Details}
We implement our models using PyTorch and MMGeneration toolkit~\cite{2021mmgeneration}. Our NeRF renderer is based on a public codebase torch-ngp~\cite{torch-ngp}, which employs a density-based grid pruning strategy for efficient real-time rendering.

\subsection{Hyperparameters}

Table~\ref{tab:params} presents the complete list of architecture/training/testing hyperparameters used in our experiments. It is worth noting that we adopt step decay policy for both the learning rate and number of inner loop iterations $K_\mathrm{in}$ during training. 

The major difference between unconditional- and reconstruction-purposed models is the training schedule, where reconstruction-purposed training stops early at 80K iterations, as mentioned in the main paper. Other differences lie in the U-Net dropout rate and latent learning rate, which may have marginal effects on the reconstruction performance.

Regarding the Langevin correction step in the form of $x^{(t)} \gets x^{(t)} - \frac{1}{2} \delta \sigma^{(t)} \hat{\epsilon} + \sqrt{\delta} \sigma^{(t)} \epsilon$ with step size $\delta$ and independent noise $\epsilon \sim \mathcal{N}(0, I)$, we observe that this technique is more effective in reconstructing Chairs than Cars. Therefore, to reduce inference time, Langevin correction is not used for SRN Cars dataset. Our intuition is that Chairs dataset exhibits higher variety in geometry, and Langevin correction helps better explore the latent space by injecting random noising during sampling.

\subsection{Training and Inference Time}

We train all our models using two RTX 3090 GPUs, each processing a batch of 8 scenes. On average, a single outer training step takes around 0.5 sec, 80K iterations take around 11 hours, and 1M iterations cost around 6 days.

Under the unconditional generation setting (50 DDIM steps), sampling a batch of 8 scenes takes 4.63 sec on a single RTX 3090 GPU. Under the reconstruction setting with the same batch size, a single guided DDIM step or Langevin step takes 0.21 sec, and a single outer finetuning step takes 0.28 sec (when $K_\mathrm{in} = 4$). This sums up to around 23 sec for reconstructing a batch of 8 Cars (single-view), and 102 sec for reconstructing a batch of 8 Chairs (single-view) with additional Langevin steps. Once the triplane latent codes are sampled, neural rendering can be performed in real time to synthesize the output images.

\section{Additional Model Details}
In the interest of reproducibility, this section provides additional details about the models used in our experiments. These techniques were not discussed in the main paper, because they are not essential components of the proposed method, and they seem to have negligible effect on the overall results (Table~\ref{tab:modeldetails}). Nevertheless, we have included them in our implementation to maintain consistency with an earlier version of our codebase where they were found to be useful at one stage.

\subsection{Bounding the Latents via Tanh Mapping}

\begin{table}[t]
\begin{center}
\scalebox{0.85}{%
\begin{tabular}{lcccc}
\toprule
Method & PSNR\textuparrow & SSIM\textuparrow & LPIPS\textdownarrow & FID\textdownarrow\\
\midrule
SSDNeRF (standard) & 23.52 & 0.913 & 0.078 & 16.39 \\
W/o Tanh & 23.59 & 0.913 & 0.077 & 16.34 \\
W/o L2 regularization & 23.48 & 0.913 & 0.077 & 16.62 \\
\bottomrule
\end{tabular}}
\end{center}
\caption{Single-view reconstruction results on SRN Cars, showing that Tanh and L2 regularization are likely to be redundant.}
\label{tab:modeldetails}
\end{table}

In an earlier version of our implementation of the diffusion model, we use the $\hat{\epsilon}$ prediction format as in DDPM~\cite{ddpm} instead of the current $\hat{v}$ format proposed by \cite{velocity}. To stabilize denoising-based sampling process, the $\hat{\epsilon}$ format requires clipping the denoised prediction $\hat{x}$ at each step, which is suitable for bounded data. This motivated us to bound the latent code $x_i$ element-wise via an additional Tanh layer. 

Specifically, let $x_i \coloneqq s \cdot \tanh{x_i^\mathrm{raw}}$ be the bounded latent code within the interval $(-s, s)$, where $x_i^\mathrm{raw}$ denotes a raw, unbounded parameterization of the code. During single-stage training and test-time finetuning, we perform optimization on the leaf variable $x_i^\mathrm{raw}$ in the unbounded space. During test-time sampling, the denoised prediction $\hat{x}$ is thus hard-clipped to $[-s, s]$ as well. We set the scale hyperparameter $s$ to 2 in all our experiments.

Because our final models have switched to the $\hat{v}$ prediction format, Tanh mapping may not be an essential component of SSDNeRF, as indicated in Table~\ref{tab:modeldetails}.

\begin{table*}[t]
\begin{center}
\scalebox{0.85}{%
\begin{tabular}{ccccccccc}
\toprule
$N_\mathrm{v}$ & View indices & $K_\mathrm{out}$ & $K_\mathrm{in}$ & LR & PSNR\textuparrow & SSIM\textuparrow & LPIPS\textdownarrow & FID\textdownarrow \\
\midrule
\phantom{0}1 & 64 & \phantom{0}25 & 4 & 0.005 & 23.52 & 0.913 & 0.078 & 16.39 \\
\phantom{0}2 & 64, 104 & \phantom{0}50 & 4 & 0.01\phantom{0} & 26.49 & 0.944 & 0.054 & 10.66 \\
\phantom{0}4 & 0, 83, 167, 250 & 100 & 4 & 0.02\phantom{0} & 28.29 & 0.955 & 0.049 & 11.09 \\
\phantom{0}8 & 0, 36, 71, 107, 143, 179, 214, 250 & 160 & 5 & 0.04\phantom{0} & 31.26 & 0.973 & 0.035 & \phantom{0}8.54 \\
16 & \scalebox{0.72}{0, 17, 33, 50, 67, 83, 100, 117, 133, 150, 167, 183, 200, 217, 233, 250} & 200 & 8 & 0.08\phantom{0} & 34.31 & 0.986 & 0.018 & \phantom{0}3.09 \\
32 & \scalebox{0.72}{\makecell{0, 8, 16, 24, 32, 40, 48, 56, 65, 73, 81, 89, 97, 105, 113, 121, \\ 129, 137, 145, 153, 161, 169, 177, 185, 194, 202, 210, 218, 226, 234, 242, 250}} & 200 & 8 & 0.08\phantom{0} & 35.66 & 0.989 & 0.015 & \phantom{0}2.35 \\
\bottomrule
\end{tabular}}
\end{center}
\caption{Details on sparse-to-dense reconstruction on SRN Cars dataset, including the number of input views $N_\mathrm{v}$ and their indices, number of finetuning outer loop iterations $K_\mathrm{out}$, number of finetuning inner loop iterations $K_\mathrm{in}$, finetuning learning rate of the latent code, and novel view synthesis evaluation results.}
\label{tab:params2}
\end{table*}

\begin{figure*}[t]
\begin{center}
\ifthenelse{\boolean{arxiv} \OR \boolean{inSubfiles}}
    {\includegraphics[width=0.68\linewidth]{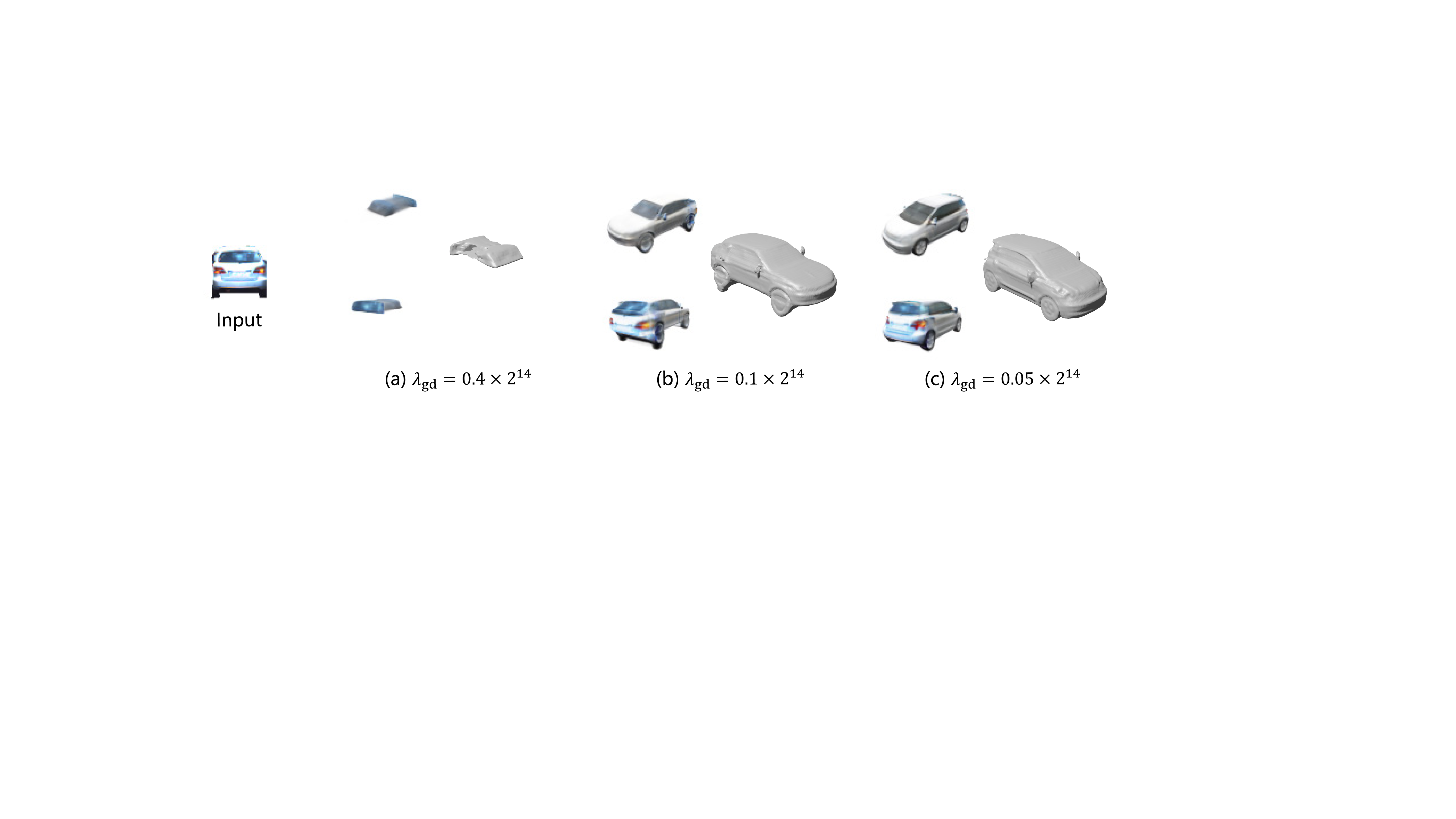}}
    {\includegraphics[width=0.65\linewidth]{media/failure.pdf}}
\end{center}
\caption{Failure case (a) and (b) in single-view NeRF reconstruction from real images. Sample (c) resolves this issue by reducing the guidance scale $\lambda_\mathrm{gd}$.}
\label{fig:failure}
\end{figure*}

\subsection{Additional L2 Regularization}

L2 latent regularization in auto-decoder training originates from the assumed Gaussian latent prior~\cite{deepsdf}. In two-stage diffusion NeRF~\cite{diffrf} or occupancy field~\cite{triplanediff} models, L2 regularization helps control the norm of the latent codes and discourage outlying values with respect to the clipping during sampling. During single-stage training and test-time finetuning, we also keep this regularization term in the actual loss function:
\begin{align}
    \mathcal{L} =\ &\lambda_\mathrm{rend}\mathcal{L}_\mathrm{rend}\left(\{x_i\},\psi\right) + \lambda_\mathrm{diff}\mathcal{L}_\mathrm{diff}\left(\{x_i\},\phi\right) \notag\\
    &+ \frac{\lambda_\mathrm{reg}}{\dim{(X)}}\expect_i{\left[\|x_i\|^2_F\right]},
    \label{actualmainloss}
\end{align}
where $\dim{(X)}$ is the latent dimensionality,
and the regularization weight $\lambda_\mathrm{reg}$ is set to $0.003$. However, as suggested in Table~\ref{tab:modeldetails}, L2 regularization also has negligible impact under the single-stage training framework.

\section{Experiment Details and Additional Results}

\subsection{Details on Sparse-to-Dense Reconstruction}

Table~\ref{tab:modeldetails} presents more details on the experiment settings, testing hyperparameters, and evaluation results of sparse-to-dense reconstruction on SRN Cars dataset. 

Overall, we find that more iterations and higher learning rate are required when finetuning on more input views, but the learning rate should not exceed the upper bound of 0.08 for stability, and a maximum of 200 outer loop iterations (totaling 1600 inner loop iterations) are sufficient for dense-view settings. 

\begin{figure*}[t]
\begin{center}
\ifthenelse{\boolean{arxiv} \OR \boolean{inSubfiles}}
    {\includegraphics[width=0.83\linewidth]{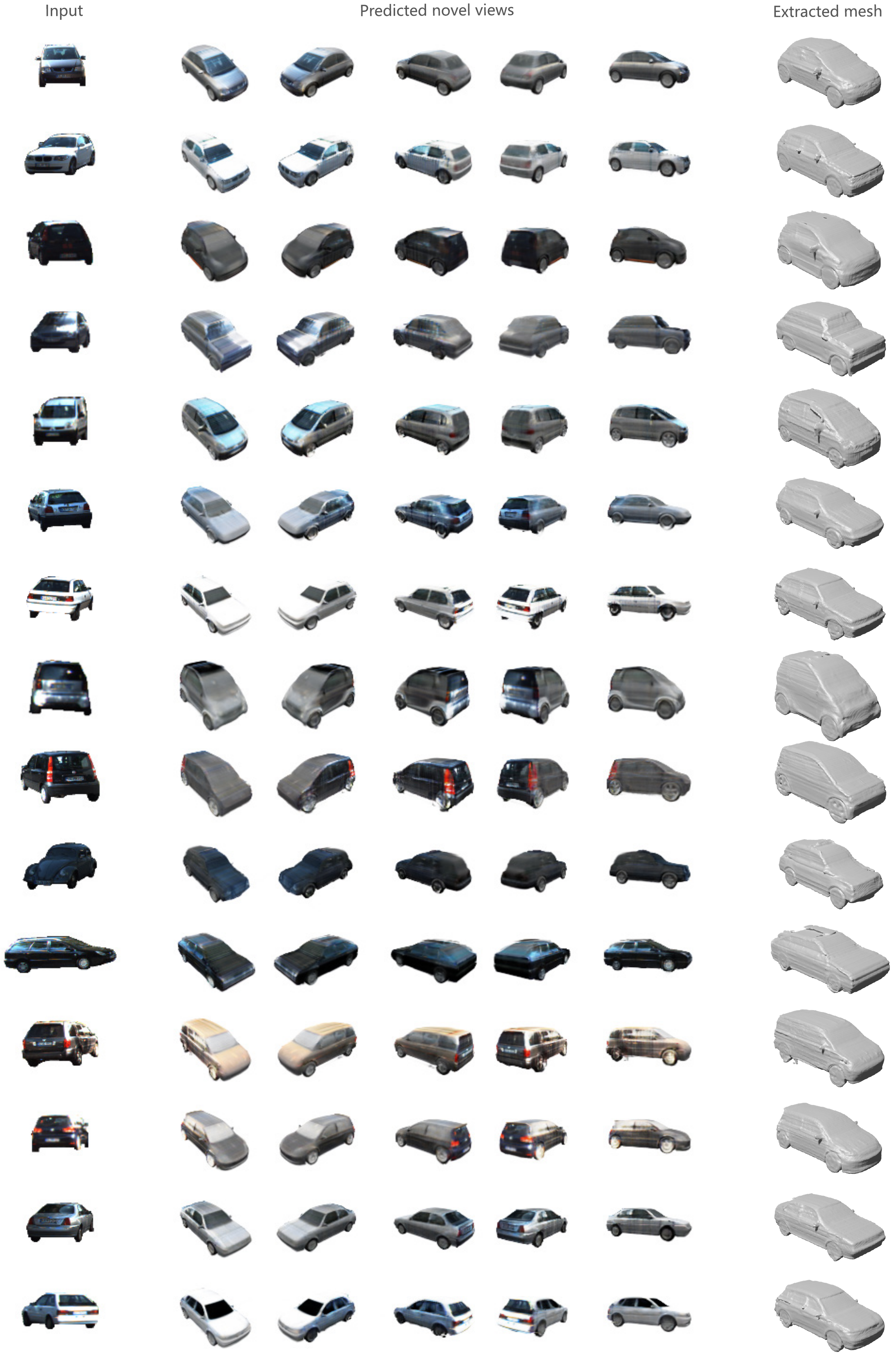}}
    {\includegraphics[width=0.72\linewidth]{media/kitti.pdf}}
\end{center}
\caption{Single-view NeRF reconstruction from real images.}
\label{fig:kitti}
\end{figure*}

\subsection{Single-View Reconstruction from Real Images}

In this subsection, we provide addition experiments on single-view NeRF reconstruction from real images, using the model trained on the synthetic SRN Cars dataset. This demonstrates the generalization capability of SSDNeRF under substantial domain gap.

\paragraph{Data Preparation} We extract images of vehicles from the KITTI 3D object detection dataset~\cite{kitti}, which provides annotated 3D bounding boxes of objects in the camera view. We use the provided ground truth bounding box dimensions and poses to align the objects in the same world coordinate system as in SRN Cars dataset. In addition, we leverage the segmentation masks annotated by Heylen~\etal~\cite{monocinis} to remove the background. All images are cropped and resized to 128\texttimes128. In real applications, one could also use a monocular 3D object detector and an instance segmentation model to obtain these inputs.

\paragraph{Testing Hyperparameters} We enable Langevin correction (5 iterations) to better handle out-of-distribution scenes, and we adopt a different setting of guidance scale $\lambda_\mathrm{gd} \coloneqq 0.4 \times 2^{14}$ and finetuning diffusion weight constant $c^\prime_\mathrm{diff} \coloneqq 4$.

\paragraph{Qualitative Results and Failure Case} 
We present qualitative examples of novel views and extracted meshes in Figure~\ref{fig:kitti}. Apart from that, we have also noticed a failure case where a large portion of the geometry is missing (Figure~\ref{fig:failure}~(a)). Nevertheless, this issue can be resolved by reducing the guidance scale $\lambda_\mathrm{gd}$ (Figure~\ref{fig:failure}~(c)). Overall, we observed that a guidance scale that is too large can result in an unstable sampling process, ultimately leading to corrupted geometries.

\subsection{Addition Qualitative Examples}

\begin{figure*}[b]
\begin{center}
\includegraphics[width=0.9\linewidth]{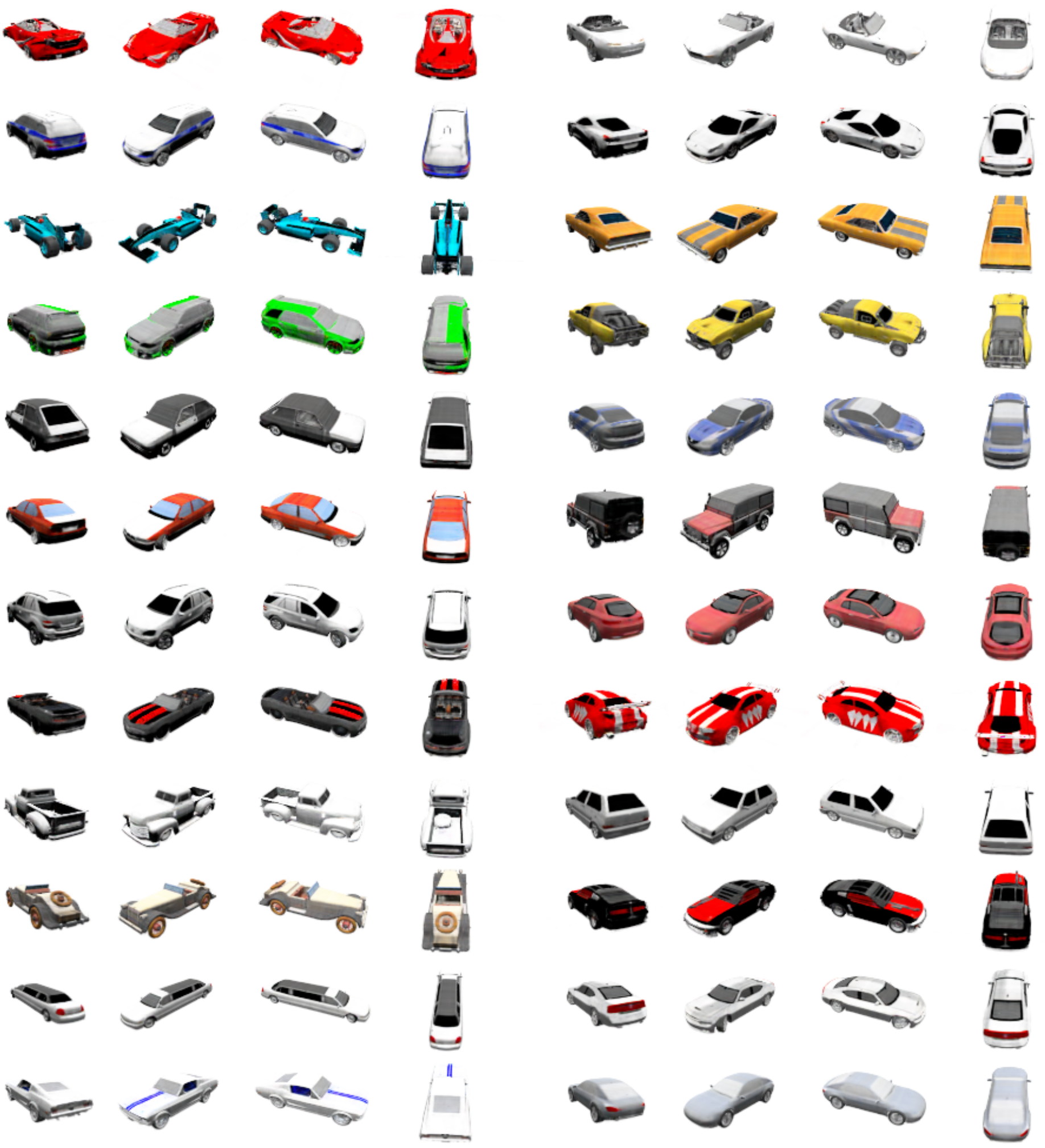}
\end{center}
\caption{Uncurated samples generated by SSDNeRF trained on SRN Cars dataset.}
\label{fig:uncond_cars}
\end{figure*}

\begin{figure*}[t]
\begin{center}
\includegraphics[width=0.88\linewidth]{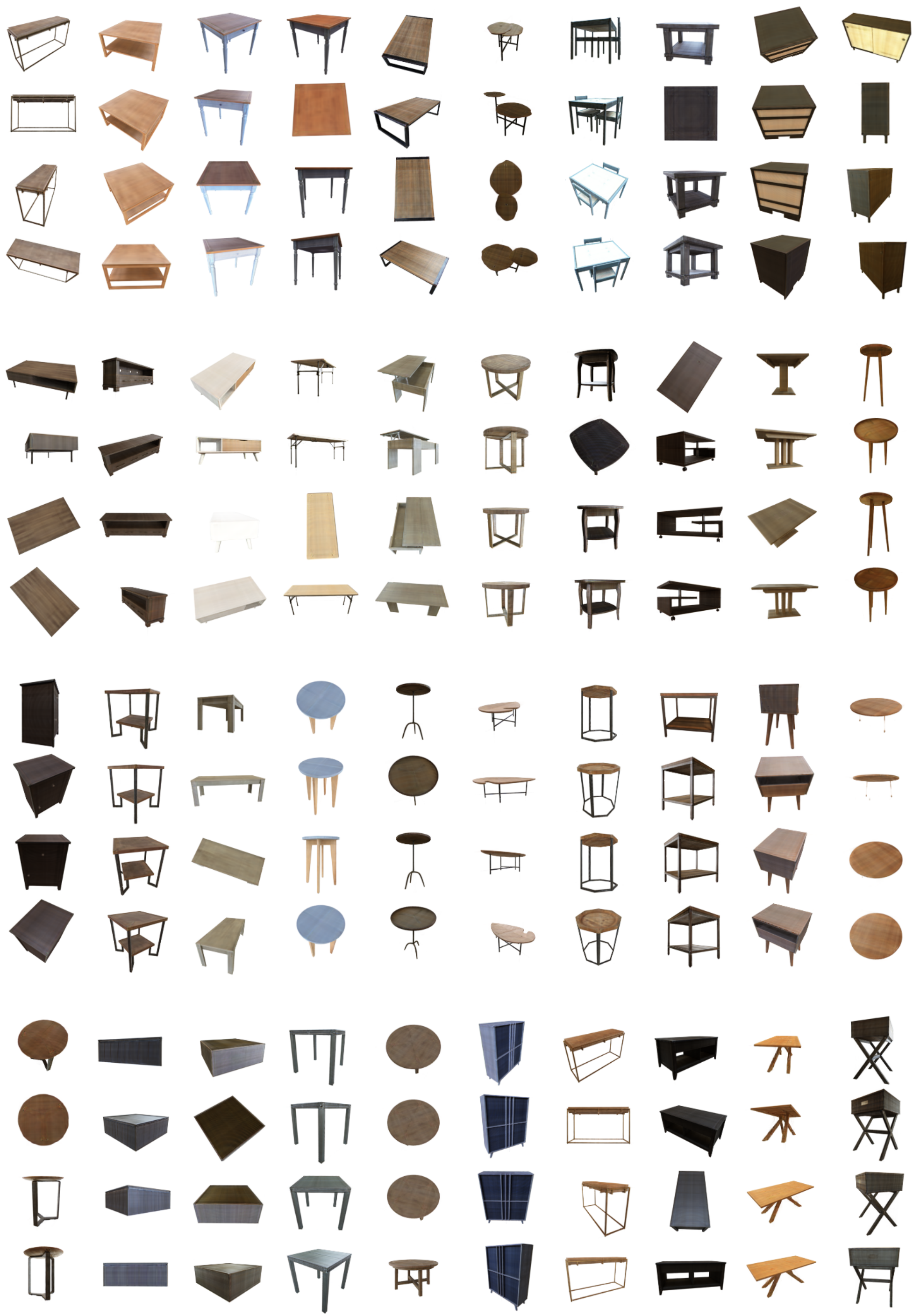}
\end{center}
\caption{Uncurated samples generated by SSDNeRF trained on ABO Tables dataset.}
\label{fig:uncond_tables}
\end{figure*}

\begin{figure*}[t]
\begin{center}
\includegraphics[width=0.9\linewidth]{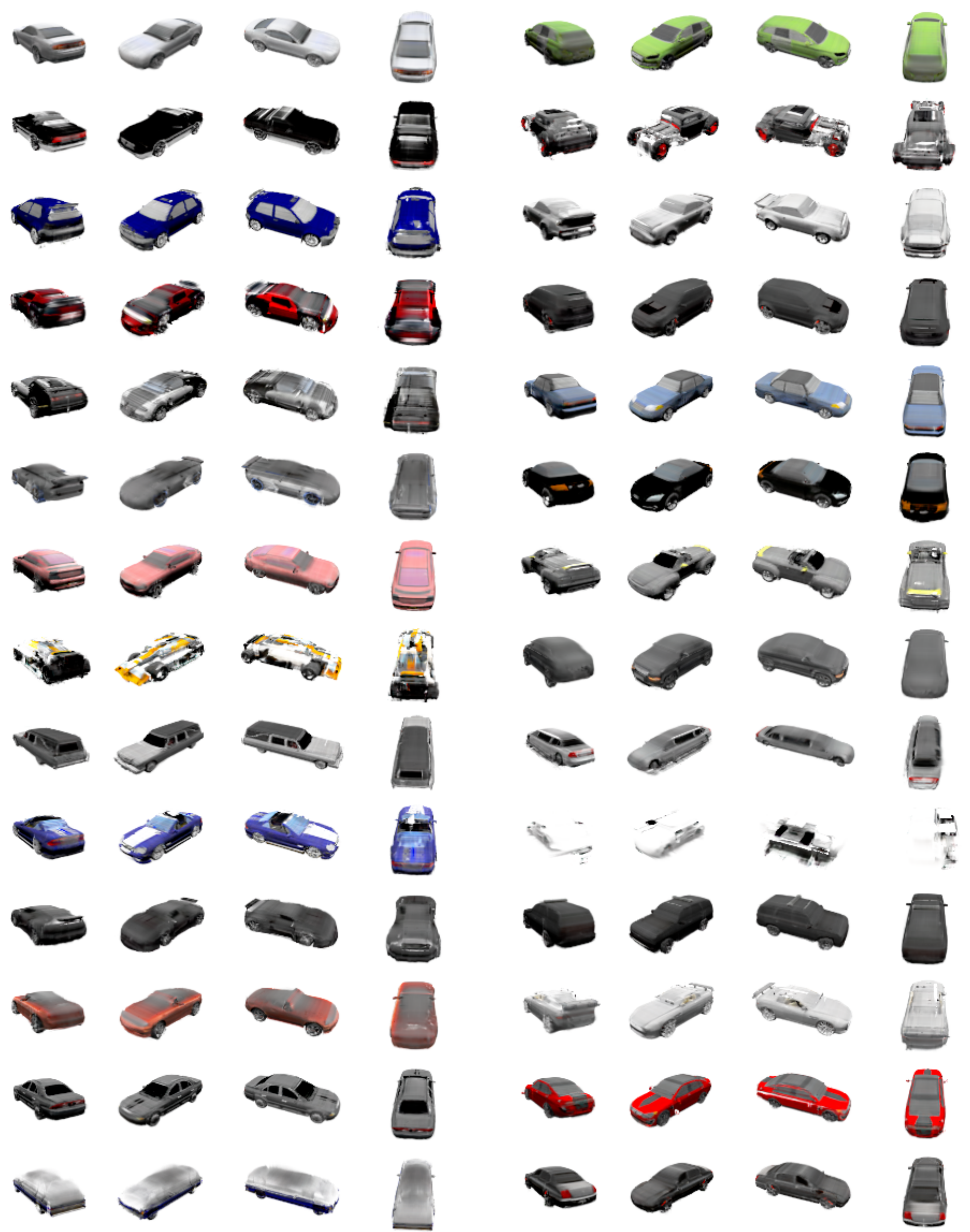}
\vspace{0.5ex}
\end{center}
\caption{Uncurated samples generated by SSDNeRF trained on a 3-view subset of SRN Cars. Note that the failure case (right column, fifth row from the bottom) is caused by the few outlier training samples, in which the objects are not properly aligned in scale and position due to a data preprocessing issue in SRN Cars~\cite{srn}.}
\label{fig:uncond_cars3v}
\end{figure*}

\begin{figure*}[t]
\begin{center}
\includegraphics[width=0.75\linewidth]{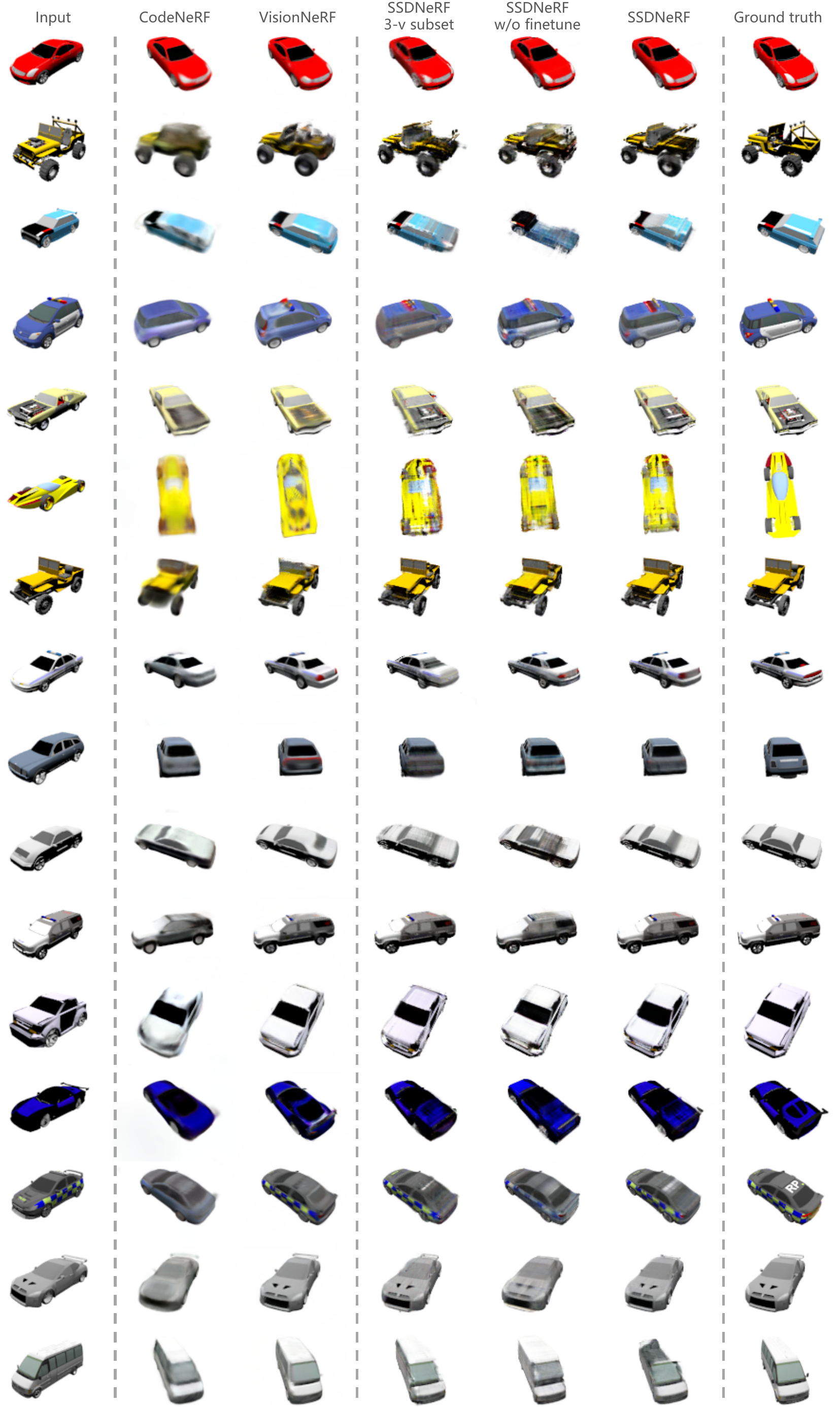}
\end{center}
\caption{Single-view reconstruction on unseen test objects in SRN Cars.}
\label{fig:cond_cars}
\end{figure*}

\begin{figure*}[t]
\begin{center}
\includegraphics[width=0.71\linewidth]{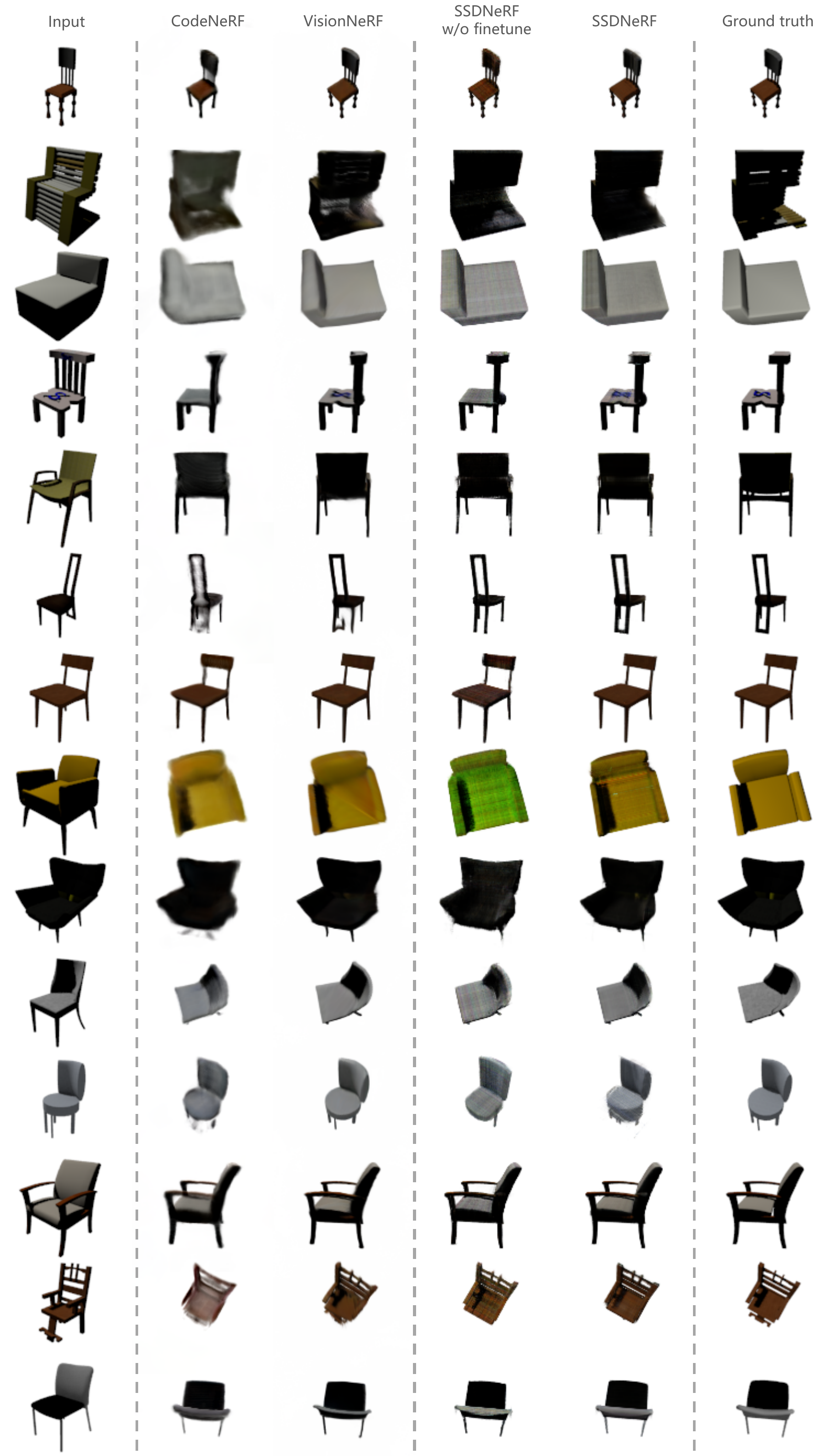}
\end{center}
\caption{Single-view reconstruction on unseen test objects in SRN Chairs.}
\label{fig:cond_chairs}
\end{figure*}

We show randomly sampled scenes generated by SSDNeRF in Figure~\ref{fig:uncond_cars}, Figure~\ref{fig:uncond_tables}, and Figure~\ref{fig:uncond_cars3v}. For single-view reconstruction, we compare the novel views predicted by SSDNeRF to those predicted by CodeNeRF~\cite{codenerf} and VisionNeRF~\cite{visionnerf} in Figure~\ref{fig:cond_cars} and Figure~\ref{fig:cond_chairs}.

\clearpage
\bib